\documentclass[10pt,twocolumn,letterpaper]{article}

\usepackage{cvpr}
\usepackage{times}

\usepackage{multirow}
\usepackage{booktabs}
\usepackage{setspace}
\usepackage{amssymb}
\usepackage[misc]{ifsym}

\usepackage{url}

\usepackage{epsfig}
\usepackage{graphicx}

\usepackage{caption}
\usepackage{amsmath}
\usepackage{amssymb}
\usepackage{cite}
\usepackage{verbatim}
\usepackage{amsthm,amsmath,amssymb}
\usepackage{mathrsfs}
\usepackage{subfigure}  

\makeatletter 
\renewcommand{\@thesubfigure}{\hskip\subfiglabelskip}
\makeatother

\usepackage{algorithm}
\usepackage{algpseudocode}
\usepackage{amsmath}
\usepackage{graphics}
\usepackage{epsfig}
\usepackage{color}

\usepackage[pagebackref=true,breaklinks=true,letterpaper=true,colorlinks,bookmarks=false]{hyperref}

\cvprfinalcopy % *** Uncomment this line for the final submission

\begin{document}

%%%%%%%%% TITLE

\title{\vspace{-66pt}Super-Resolving Compressed Video in Coding Chain}

\author{
	Dewang Hou$^{1}$ \quad Yang Zhao$^{2,3}$  \quad Yuyao Ye$^{1}$ \quad Jiayu Yang$^{1}$ \\ \\
	Jian Zhang$^{1, 3}$  \quad Ronggang Wang$^{1, 3}  $  \\
\\
	$^1$Peking University \quad $^2$ Hefei University of Technology \quad
	$^3$Peng Cheng Laboratory\\
	
	\vspace{-0.5em}
}

\twocolumn[{%
	\renewcommand\twocolumn[1][]{#1}%
	\maketitle
	% Remove page # from the first page of camera-ready.
	% \ificcvfinal\thispagestyle{empty}\fi
	%\thispagestyle{empty}

\begin{center}
	\centering
	\includegraphics[width=1\textwidth]{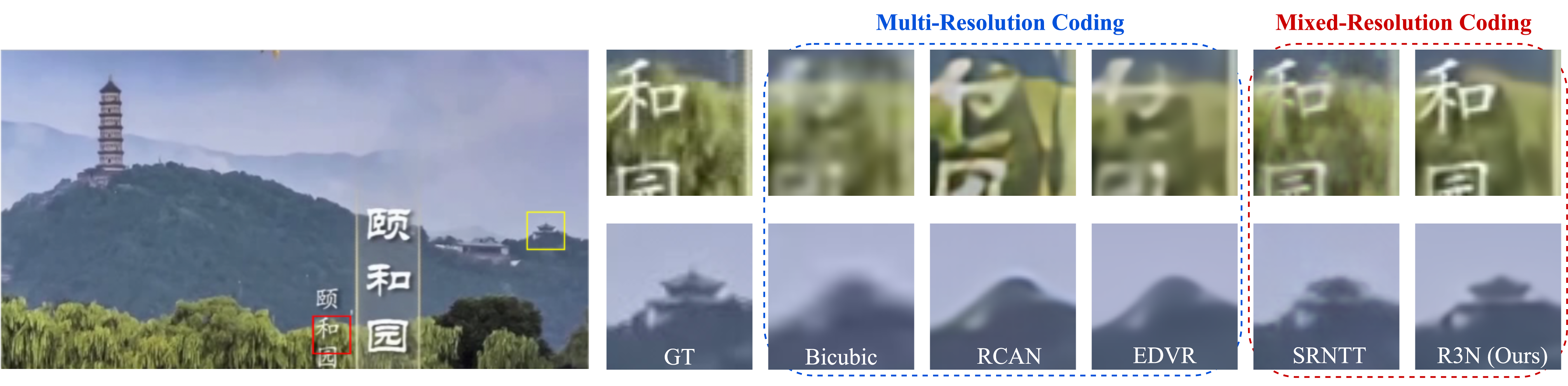}

	\captionof{figure}{
		A comparison of visual quality on low-resolution compressed video restoration (reconstructing 1080p results from  270p compressed videos). 
		RCAN \cite{zhang2018image}, EDVR \cite{wang2019edvr}, SRNTT \cite{zhang2019image} are the state-of-the-art methods for single image, video and  reference-based super-resolution, respectively.
		In this paper, bicubic, RCAN and EDVR are used in the multi-resolution coding pipeline; SRNTT and the proposed R3N are used in the mixed-resolution coding pipeline.
	}
	\label{fig_result}
	\vspace{12pt}
\end{center}
}]

%%%%%%%%% ABSTRACT
\begin{abstract}
	Scaling and lossy coding are widely used in video transmission and storage.
	Previous methods for enhancing the resolution of such videos often ignore the inherent interference between resolution loss and compression artifacts, which compromises perceptual video quality. To address this problem, we present a mixed-resolution coding framework, which cooperates with a reference-based DCNN.
	In this novel coding chain, the reference-based DCNN learns the direct mapping from low-resolution (LR) compressed video to their high-resolution (HR) clean version at the decoder side. 
	We further improve reconstruction quality by devising an efficient deformable alignment module with receptive field block to handle various motion distances and introducing a disentangled loss that helps networks distinguish the artifact patterns from texture. Extensive experiments demonstrate the effectiveness of proposed innovations by comparing with state-of-the-art single image, video and reference-based restoration methods.

\end{abstract}

%%%%%%%%% BODY TEXT
\section{Introduction}

Nowadays, video commands the lion’s share of Internet traffic and is still climbing. 
On the one hand, it is continuously desirable to develop coding algorithms with a higher compression ratio \cite{sullivan2012overview} and lower complexity.
On the other hand, technically, a post-processing step of restoring videos at the decoder side has more room for improvement compared with the standards.

To achieve optimal video quality under bandwidth and power constraints, one effective way is to down-sample the video before compression and transmission.
Therefore, resolution loss and compression artifacts are the most typical examples of degradation in practical applications. For the task of recovering a clean video from its degraded version at the decoder side, we call it restoration strategy decompression.
It is a highly ill-posed problem as there exists infinite feasible solutions. 

Fortunately, although super-resolution (SR) and compression artifact removal (CAR) are long-standing tasks, they are progressing rapidly driven by machine learning advances. Especially, convolutional neural networks \cite{dong2014learning, dong2015compression} have been successfully utilized for low-level vision tasks.
However, the majority of previous research has focused on single well-defined problem. 
On the contrary, joint SR-CAR is an intricate task, where high-frequency details are restored for SR and high-frequency artifact are removed for CAR.
We will elaborate it in Sec.~\ref{sec2.3}.

Thus, when it comes to practical applications for joint SR-CAR, the difficulty is increased and less progress has been made. 
To make a step forward, we rethink the coding framework \cite{schwarz2007overview, boyce2015overview, maurer2020overview} and post-processing from a holistic perspective. The key innovation is a novel paradigm that can leverage synergies in coding and post-processing. 
Specifically, we propose a modified framework, Mixed-Resolution Coding (MixedRC). The term “mixed-resolution” in this paper refers to dual bitstream, one of which contains high-frequency information from full-resolution key-frames and the other contains information from reduced-resolution frames. 
And we design a neural network architechture, called Reference-based Restoration Network with Refined-offset deformable alignment (R3N), to relieve performance bottlenecks of ill-posed problems.
A trained R3N can efficiently transfer the textures from key-frames to LR compressed frames and achieve highly attractive performance. We illustrate comparison examples in Fig.~\ref{fig_result}.
Last but not least, we  improve reconstruction quality by introducing a disentangled loss that classifies pixels into texture and artifact.

In summary, the main contributions are three-fold:

\begin{itemize}
    \item We propose MixedRC,  which can be seamlessly incorporated into existing (and future) single-layer codecs to enhance their performances with minimal effort. This study also breaks new ground for reference-based networks in post-processing.
   	\item  We define a disentangled loss that helps the network distinguish the artifact patterns from texture by comparing statistics collected over the entire image. Ablation experiments further verify the effectiveness of the proposed innovations.
	\item We propose a reference-based network, called R3N. The core of R3N is the refined-offset deformable alignment, which is efficient to implement and is easy to optimize. 
	Receptive field block and spatial attention in this module can improve performance by “calibrating” feature responses, which is a far more flexible alternative to handle various motion distances.  And we reveal that a easy-to-hard transfer, setting on frame rate, is helpful in learning a reference-based model.

\end{itemize}

\begin{figure*}
	\centering
	\subfigure[(a) Conventional Coding chain]{
		\begin{minipage}[b]{1\linewidth}
			\includegraphics[width=1\linewidth]{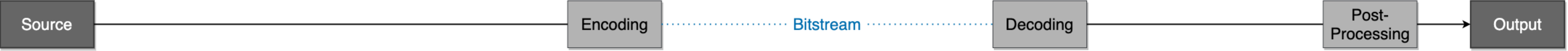}
			\vspace{-12pt}
	\end{minipage}}
	\subfigure[(b) MultiRC chain]{
		\begin{minipage}[b]{1\linewidth}
			\includegraphics[width=1\linewidth]{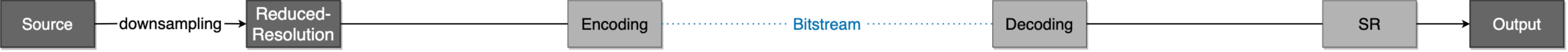}
			\vspace{-12pt}
	\end{minipage}}
	\subfigure[(c) MixedRC chain]{
		\begin{minipage}[b]{1\linewidth}
			\includegraphics[width=1\linewidth]{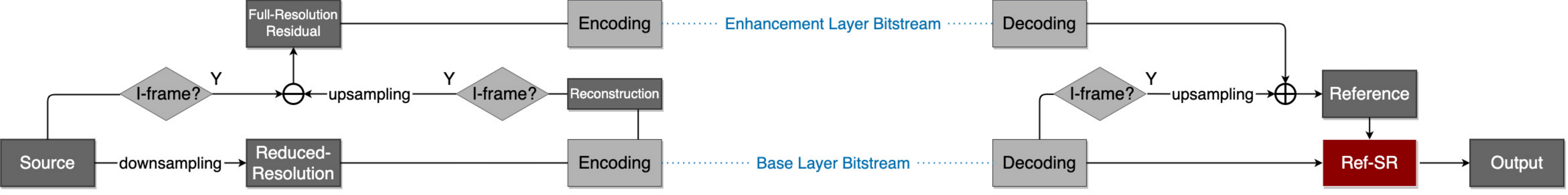}
			\vspace{-12pt}
	\end{minipage}}
	\vspace{-6pt}
	\caption{The scheme of mentioned coding frameworks. In MixedRC, frames scaled to different resolutions are encoded into layered bitstreams.}
	\label{fig_coding}
	\vspace{-10pt}
\end{figure*}

% $\bullet$ 

% relatedworkrelatedworkrelatedworkrelatedworkrelatedworkrelatedworkrelatedworkrelatedworkrelatedworkrelatedworkrelatedworkrelatedworkrelatedwork
\section{Related Works}
\subsection{Multi-Resolution Coding.}

It is known that, at low bit rates, a down-sampled video visually beats the HR video when represented with the same number of bits via compression.
In 2003, Bruckstein \textit{et al.} gave a numerical analysis of the down-sampling and compression process \cite{bruckstein2003down}. 
After that, studies \cite{shen2011down, georgis2015reduced, fischer2020versatile} combined video coding and SR techniques to improve coding efficiency.  

Among them, Multi-Resolution Coding (MultiRC) \cite{holcomb2008multi} adaptively changes frame sizes for artifact and complexity reduction. However, the limited quality of frames, with no exception, narrows down the scope of this framework and becomes harder to super-resolve frames with artifact removal jointly, in terms of expressiveness and robustness \cite{yu2018crafting, zhang2018learning}.
The rudiment of MixedRC is also proposed in that decade \cite{brandi2008super}. 
However, limited by the performance of SR at the time, these pioneering works have not been well developed. 
To advance this line, we propose a MixedRC based on scalable system \cite{schwarz2007overview, maurer2020overview}. Both MultiRC and the proposed MixedRC are elaborated in Fig.~\ref{fig_coding}.

\subsection{ Super-Resolution}
\noindent{\bfseries Loss functions for SR.}
Sophisticated SR algorithms often focused on minimizing pixel-wise reconstruction errors to achieve a high peak signal-to-noise ratio (PSNR), which correlate poorly with image quality as perceived by a human observer \cite{wang2009mean}. 
Parallel efforts also studied adopting perceptually-motivated losses \cite{mathieu2015deep, zhao2016loss, johnson2016perceptual} to avoid the pixel-wise average problem, which typically leads to over-smoothed results.
Orthogonal to these are adversarial losses \cite{ledig2017photo, wang2018esrgan}, pushing the super-resolved image to be of high likelihood given examples from HR domains.

\noindent{\bfseries Reference-based SR.}
The lost information during down-sampling an HR image amounts to high-frequency components \cite{shannon1949communication}. Intuitively, to restore such high-frequency components, providing similar rich textures is a more reasonable approach than generating unreal textures. Early manually-designed filters \cite{he2012guided, ham2015robust} use an external image as guidance to adjust filter parameters, which can preserve sharp edges. 
These works have seminal significance on reference-based SR (Ref-SR) \cite{zheng2018crossnet, zhang2019image, shim2020robust}, which aims to super-resolve an LR image with the help of HR reference.
The main challenge lies in the design of the alignment module: CrossNet \cite{zheng2018crossnet} performs alignment based on optical flow estimations and warping operations. However, flow estimations \cite{ilg2017flownet} need additional supervision and are highly vulnerable to large motions. Later, SRNTT \cite{zhang2019image} relaxes constraint on content similarity levels by adopting brute-force patch matching at multi-scale feature space. Along this line of thought, Zhang \textit{et al.}\cite{zhangtexture} achieved significantly improved results on a challenging task of super-resolving painting images.
Shim \textit{et al.} \cite{shim2020robust} pointed out that deformable convolution \cite{dai2017deformable} is a preferred mechanism to perform implicit motion compensation. 
It is worth noting that the difficulty of alignment in the Ref-SR is more incredible than adjacent frame alignment in video SR \cite{wang2019edvr, tian2020tdan}. To this end, Shim \textit{et al.} use the stacking of deformable convolution layers to sample more locations with a larger receptive field. However, the deformable convolution is usually tricky to train, let alone stacking multiple layers sequentially. Moreover, stacking layers ``plainly" could be suboptimal for sampling locations at long distances.

% CARCARCARCARCARCARCARCARCARCARCARCARCARCARCARCARCARCARCARCARCARCARCARCARCARCARCARCARCARCARCARCARCARCARCARCARCARCARCARCARCARCARCARCAR
\subsection{Compression Artifact Removal}
CAR is also a long-standing low-level vision task. Compared to image CAR, learning-based video CAR appears as a mostly under-explored field. There is less complementary information between adjacent compressed frames due to the nature of coding algorithms, that quantization is enforced on the prediction error during inter-frame encoding. Moreover, the positions where blocking boundaries appear are relatively changeless in adjacent frames. 
% Dangling modifier ??
To circumvent this problem, methods tend to extract temporal information that lies in the dependencies between the current frame and some valuable frames, such as peak quality frames \cite{2019MFQE} and previous restored frames \cite{lu2018deep}. While in our algorithm, these valuable frames are specifically Ref frames. 

\subsection{Restoration of Multiple Degradations}
\label{sec2.3}

Scaling and compression cause the dominant degradations during video transmission.
A network that can juggle SR and CAR  tasks jointly is far more desirable than cascading different networks. 

Recent studies \cite{kim2019deep, xiang2020zooming, liu2020joint, qian2019trinity, suganuma2019attention} show the advantages of the joint-learning on multiple-degradation problem. 
Yu \textit{et al.} \cite{yu2018crafting} observed that the restoration of multiple degradations is not a simple composition of corresponding restorers trained on specific tasks.
Unlike the rising interest in other joint-learning studies, the research devoted to restore the quality and practicality of LR compressed video is rarely mentioned.
More relevant to this work, Zhang \textit{et al.} \cite{zhang2018learning} proposed SRMD, a single network to perform SR and denoising jointly. 
These couple of tasks would interfere with each other, which leads to visually unpleasant results.
More specifically, directly super-resolving the noisy input will exacerbate the unwanted noise, rendering them visually objectionable. And simply pre-denoising also tends to lose details, result in worse SR performance.
SRMD can achieve satisfactory performance if the predicted degradation maps are close to the ground truth. However, it is still not applicable in real applications as the blur kernel and noise level cannot be predicted for every image on hand. 

\begin{figure*}[ht]
	\centering
	\includegraphics[width=1\textwidth]{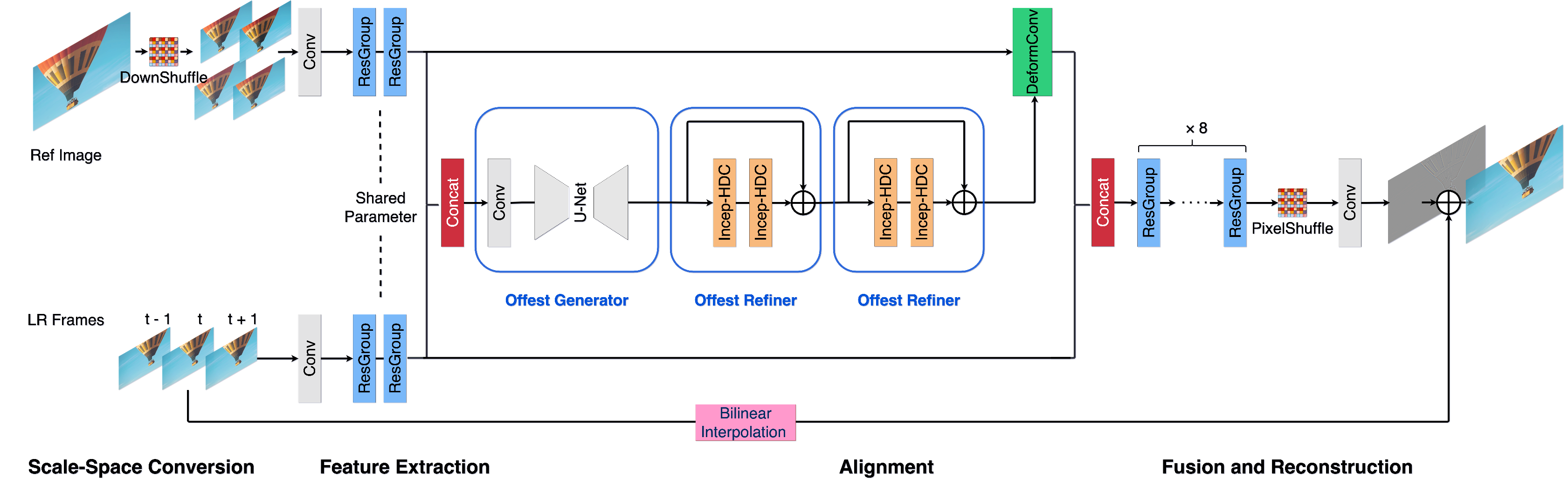}
	\vspace{-10pt}
	\caption{Network architecture of R3N. 
		It is a pipeline that consists of scale-space conversion, feature extraction, alignment, fusion and reconstruction. we adopt the \textit{ResGroups} proposed in RCAN \cite{zhang2018image} to the feature extraction and reconstruction parts.
		And for simplicity, we only show the architecture for three LR frames and one Ref image as input.}
	\label{fig_R3N}
	\vspace{-10pt}
\end{figure*}

\section{Methodology}

We first introduce the proposed MixedRC, a new coding chain based on scalable  system. 
Then for joint SR-CAR problem, we carry on simplification and improvement in a reference-based network architecture, called R3N. 
Finally, we discuss how disentangled loss function works to suit the remedy to the case of multiple-degradation problem.

%-------------------------------------------------------------------------
\subsection{Coding Framework}

The overall pipeline of the proposed coding framework is shown in Fig.~\ref{fig_coding}. 
Used by the proposed framework, spatial scalability \cite{schwarz2007overview} describes cases in which the base layer (BL) is used to encode a lower resolution signals of the video stream, and the enhancement layers (ELs) are used to encode high-frequency information in key-frames.

Another point is that video encoders divide up a video into sets of frames called group-of-pictures (GOP), specifies the order in which intra- and inter-frames are arranged. And in the proposed framework, only Intra-coded (I-) frames exist in the ELs. In addition to lower complexity, it further reduces the bit rate and the syntax overhead for enhancement layers. 
Finally, the I-frame decoded by layered coding in a GOP will be fed to a Ref-SR model as reference, to super-resolve other frames in the same GOP.
And for simplicity, backward or bi-directional prediction is not used in this paper.
MixedRC is not meant to be an alternative to existing codecs, but rather a useful complement to any codec.

\subsection{Network Architecture}

In the proposed coding framework, R3N works as a post-processing tool at the decoder side, which helps extract details lost in LR features but existed in Ref  features.
The overall structure of R3N is shown in Fig.~\ref{fig_R3N}. 
We formulate reference-based video restoration as an integrative process of transferring plausible textures conditioned on Ref images $I^{Ref}$, and at last restoring the middle frame $I^{LR}_{t}$ among $2N+1$ consecutive LR compressed frames $I^{LR}_{[t-N:t+N]}$. 

Next, we detail the individual components of the proposed R3N.

\noindent{\bfseries Down-Shuffle.} 
Scale-conversion meant to make the contents of Ref image and LR frames be on the same scale. 
In contrast to the previous works, which usually  use strided convolution or interpolation, we apply down-shuffle to accomplish scale-conversion. PixelShuffle \cite{shi2016real} is proposed for image SR, and it is also the latest up-sampling scheme used in the state-of-the-art. 
We use its inverse process, \textit{i.e.}, down-shuffle, to perform lightweight scale-space conversion. 
The down-shuffle operation systematically rearranges spatial pixels into channels, keeping the high-frequency amount intact, hence providing sufficient information for the following convolutional layers.
Note that down-shuffle is parameter-free, and it can naturally handle geometric transformations through the combinations of multiple channels with shifted features. And it turns out that down-shuffle is simple but enough to achieve competitive performance without extra cost.

\noindent{\bfseries Refined-offset Deformable Alignment.}
Deformable convolution has recently shown compelling performance in aligning frames.
Furthermore, Chan \textit{et al.} \cite{chan2020understanding} shed light on the underlying mechanism of deformable alignment, suggesting that the increased diversity in deformable alignment promises better restoration performance than flow-based alignment. 
The key idea of deformable alignment is to displace the sampling locations of standard convolution by some learned offsets. 
Through experiments, we observed that sampling more locations is not as easy as stacking more layers, which can cause instability problems of learned offsets. 

To this end, we propose a refined-offset deformable alignment to ease the training of deformable convolutional layer and steadily expand the receptive field to sample long-range locations, whose formulation is 
\begin{equation} 
	O^{1}=G(F^{Ref}, F^{LQ}),
\end{equation}
\begin{equation} 
	O^{r}=R(O^{r-1})+O^{r-1}, 2\leq r \leq n,
	\label{eq:rl}
\end{equation} 
where $G$ refers to the offset generator and $R$ refers to the offset refiner, as shown in Fig.~\ref{fig_R3N}. $F^{Ref}$ and $F^{LQ}$ refer to features of reference and LR compressed frames respectively.
Precisely, the core alignment module consists of $n$ offset refiners such that the $\left (r-1\right)^{th}$ updates the offset residues between the estimated and the previous to obtain offsets $O^{r}$. 
The rationale behind this algorithm design is residual learning\cite{he2016deep}.
We hypothesize that it is easier to update the offset residues than to estimate unreferenced offsets. 
To the extreme, if the previous offsets were optimal, it would be easier to push the residues to zero, compared with stacking deformable convolution layers, which regresses to learn ``ordinary" convolutional layers and further performs identity mapping. 
Through these benefits, refined-offset deformable alignment can handle larger motion with a wider receptive field and its coarse-to-fine strategy.

Here, layers built for residual learning also consider the eccentricity of  receptive fields. 
To enable the offset to cover a wide range of areas, sampling locations from near to far distances, we introduce a basic module, termed Inception Hybrid Dilated Convolution (Incep-HDC). 

\begin{figure}
	\centering
	\subfigure[(a) Incep-HDC]{
		\begin{minipage}[b]{0.50\linewidth}
			\includegraphics[width=0.9\linewidth, height=1.66in]{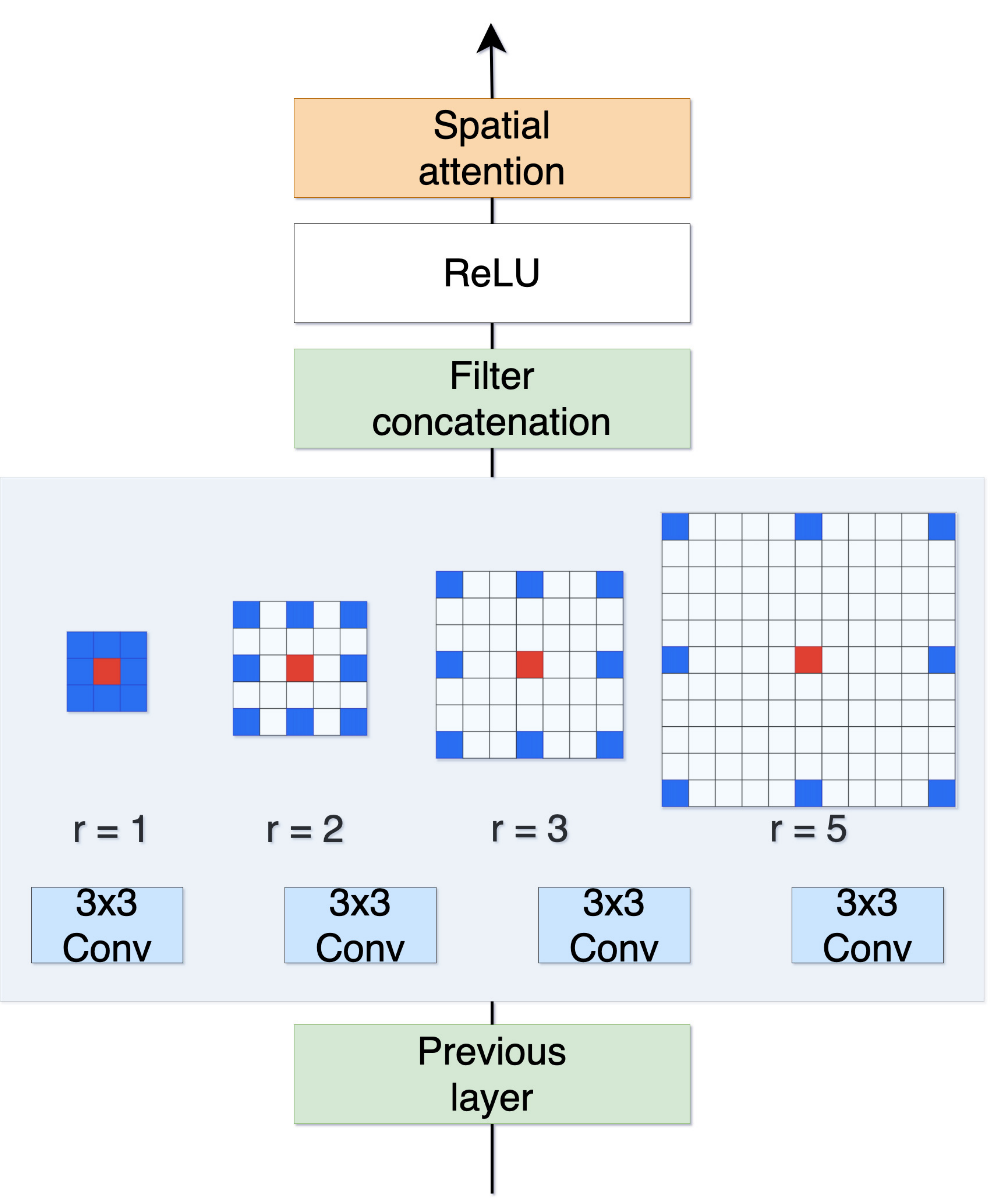}
	\end{minipage}}\hspace{0.03\linewidth}
	\label{subfig:Incep-HDC}
	\subfigure[(b) Attention Module]{
		\begin{minipage}[b]{0.29\linewidth}
			\includegraphics[width=0.9\linewidth, height=1.66in]{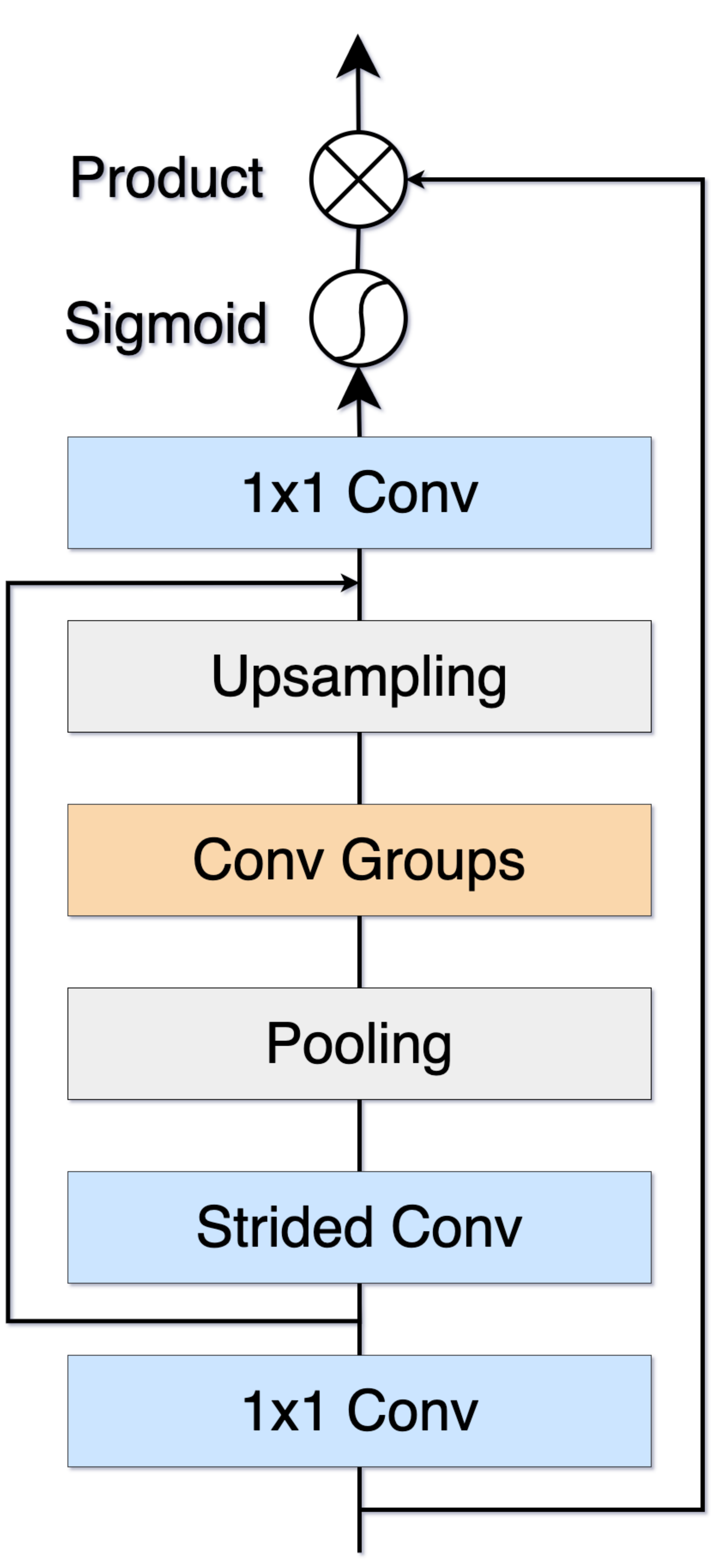}
	\end{minipage}}
	\vspace{-8pt}
	\caption{Illustration of Incep-HDC and spatial attention module.}
	\vspace{-15pt}%可以减小与下文的间隔
	\label{fig_HDC}
\end{figure}

\noindent{\bfseries Incep-HDC.}
As shown in Fig.~\ref{fig_HDC}, The inner structure of Incep-HDC can be divided into two components: the multi-branch convolution layers with different dilation rates and the spatial attention module. 

Like seminal works in semantic segmentation and object detection\cite{wang2018understanding, liu2018receptive}, recent advances in video SR\cite{isobe2020video} also carefully design the combinations of different dilation rate, according to the motion distance in divided temporal groups.
However, since both large and small motion may occur in different region of the same frame, we are motivated to utilize spatial attention mechanism.  We conduct investigation between channel attention and spatial attention, which will be detailed in Table~\ref{tab_ablation}. And we come to the conclusion that, to reach various and distant positions, multi-branch convolutions with different receptive fields are best to be used in conjunction with the spatial attention layer.

% The structure design of Incep-HDC aims to reach various and distant positions. 

\subsection{Disentangled Loss}
A trained R3N has also carried on the synthetical improvement at the aspect of loss function design.

When encountering severe noise/artifact interference, SR algorithms will suffer a huge performance degradation.  
There is a strong motivation on avoiding or mitigating such problems.
For existing methods, ranging from loop filters to data-driven algorithms, there is no mechanism or formulization to suppress various noises in signals.
Specifically, algorithms could be either too aggressive and amplify erroneous high-frequency components, or too conservative and tend to smooth over ambiguous components, both resulting in bad cases that seriously affect subjective visual impression.
To alleviate such problems, we turn to additionally introduce a novel loss function into network optimization.

\begin{figure}[t]
	\centering
	\includegraphics[width=0.485\textwidth]{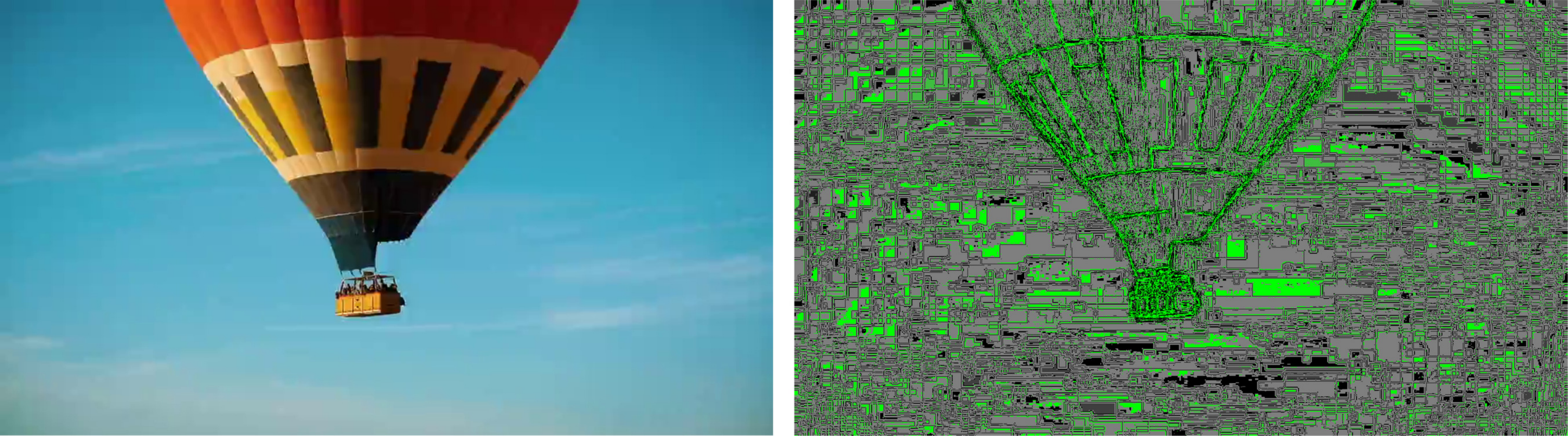}
	\caption{\textbf{Left}: A frame captured from a compressed video.  \textbf{Right}: Visualization of an analysis map on the left image. 
		The mixture of artifact, texture and flat regions are partitioned into black, green and gray pixels respectively. For instance, the ringing and motion prediction error near the boundaries that look like “mosquitos” flying around the ballon.
	}
	\vspace{-16pt}%可以减小与下文的间隔
	\label{fig_loss}
\end{figure}

\noindent{\bfseries Definition:}
Returning to the design of disentangled loss, the main idea is to introduce the texture analysis into network optimization by measuring statistics collected over the entire image. 
The disentangled loss function between the ground truth image $Y$, and the prediction $\hat{Y} = F_{SR}(X)$ is given by

\begin{equation} 
	\begin{aligned}
		\mathcal{L}_{distan}(X,\hat{Y},Y) 
		&= \ell_{p}(D(Y, X), D(\hat{Y}, X)) \\
		&= \left \| D(Y, X) - D(\hat{Y}, X) \right \|_{p},
		\label{eq:loss}
	\end{aligned}
\end{equation}
where $D(\cdot)$ is the texture analysis procedure that distinguishes the pattern of signal and noise. The proposed method tend to disentangle high frequency components correctly by minimizing a distance, for instance $\ell_{p}$ with p = 1 or p = 2. 
We next formulate $D(\cdot)$ step by step mathematically. 
Let $\uparrow$ denotes applying the bicubic upsampler, and $F_{lp}$ denotes low-pass filtering, which is implemented in a differentiable manner with \textit{Kornia} \cite{riba2020kornia}.  we derive the following procedure:

\begin{equation} 
	\mathcal{R}_{lp} = F_{lp}(x_{\uparrow}) - x_{\uparrow},
	\label{eq:a2}
\end{equation}
\begin{equation} 
	\mathcal{R}_{y} = y - x_{\uparrow}, 
	\label{eq:a1}
\end{equation}
\begin{equation} 
	% \mathcal{M} = sgn(\mathcal{R}_{y} \odot \mathcal{R}_{lp}),
	D(y, x) = sgn(\mathcal{R}_{y} \odot \mathcal{R}_{lp}),
	\label{eq:a3}
\end{equation}
where $sgn(\cdot)$ denotes signum function that extracts the sign of a given pixel value; $\odot$ is the element-wise product. 
Finally, Eq.~(\ref{eq:a3}) gives pixel-wise classification results discretized into values of -1, 0 and 1.
To facilitate understanding, we visualize an analysis result in Fig.~\ref{fig_loss}. 

\section{Dataset}
\noindent{\bfseries VSRE-set.} Youku-VSRE \cite{VSRE} collects 1,000 HD visual lossless videos from \textit{Youku} media database, and this dataset contains diverse contents that cover a variety of categories. 
It meets requirement of research on video transmission and post-processing algorithms. We reprocessed the source video from VSRE-set, and contribute a large-scale dataset for processing LR compressed video with combinations of encoding parameters, along with reference and ground truth. 
So that it can be used to train and benchmark SR on compressed videos. 

\noindent{\bfseries Our distortions.} The degradation methods of VSRE-set are described as following. 
We down-sampled HD video clips from VSRE-set with bicubic degradation on scaling factors of 2× and 4×.
To obtain the target degraded clips, we generated the compressed frames through two coding settings, i.e., the latest HEVC \cite{sullivan2012overview} standard using HM 16.0 with the Low Delay P (LDP) mode. And the quantization parameter qp = 37 on frames with scaling factors of 2×, qp = 28 on frames with scaling factors of 4×.
The above processing has constituted sets for 2× and 4× SR respectively. 
In addtion, we generated the Ref images through  SHM 6.1 with the All Intra (AI) mode at qp = 28.  
In short, each data sample consists of: source video clip, LR compressed video clip and Intra-coded video clip. LR video clip is the input and source video clip is the ground-truth. 

For training Ref-SR models, the Ref image can be choosen from the Intra-coded video clip. 
And here for testing Ref-SR, the Ref image always refers to the first frame of the Intra-coded video clip. 

\noindent{\bfseries Test set.} 
In VSRE-set, 900 video clips with a total of 90, 000 frames are used as  training set.
The remaining video clips are re-grouped, and can be used as the validation and test set.
Among them, we select 40 representative clips into our test set. 
We additionally cover more tests on 60 videos from JCT-VC \cite{bossen2013common} and VideoSet \cite{wang2017videoset}.
Finally, test set consists of 100 videos, denoted by \textit{Video-Transmission-100 (VT-100)}.
And HEVC sequences (Class A, B, E) have also been compressed with a wide range of QP spans for a wider test.

Note that, it is difficult to collect a dataset with strict bitrate control and alignment. 
The reprocessed VSRE-set mainly aims to simulate the real and complex degradation in video transmission, and to train and verify the restoration algorithms. 
However, we controlled the bitrate to the same range for experiments in Sec.~\ref{chain}.
Experiments have shown that a R3N trained on VSRE-set can effectively be generalized to other datasets.

%------------------------------------------------------------------------
\section{Experiments}
\label{sec:Experiments}

\subsection{Coding Setup and Performance}
\label{chain}
%包括Implementation details和Data和Training
For our simulation framework, the HEVC base layer was encoded by software HM 16.0 while the HEVC enhancement
layer was encoded by SHM 6.1, and Bidirectionally predicted frames (B-frames) are disabled in codec.
In practical, GOP typically contains between 6 to 16 frames that are visually similar with no scene transitions. We also set the GOP with 16 frames, so the proportion of I-frames to the total number of frames is 1/16. 
And we conduct experiments compared with the conventional and the MultiRC coding chain. Their pipelines are shown schematically in Fig.~\ref{fig_coding}. 

\begin{figure}[]
	\centering
	\includegraphics[width=0.488\textwidth]{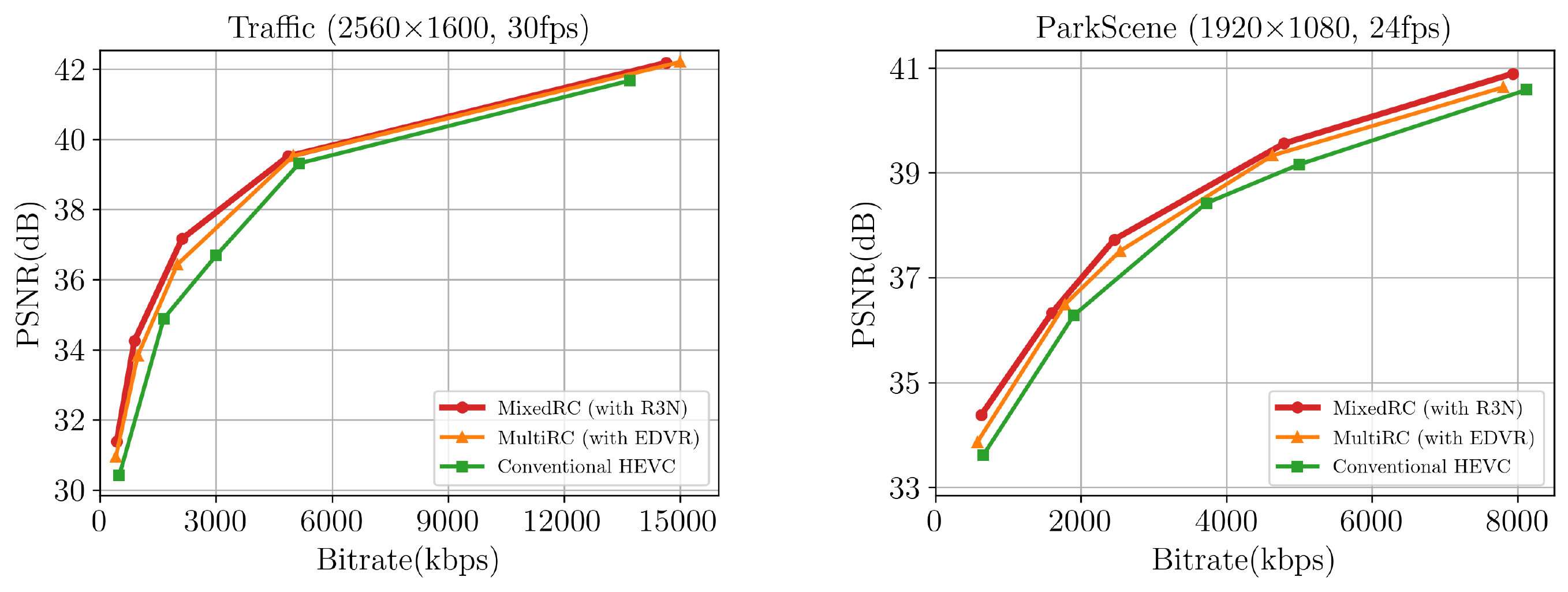}
	\vspace{-16pt}
	\caption{PSNR over bitrate for \textit{Traffic} and \textit{ParkScene}.}
	\vspace{-3pt}%可以减小与下文的间隔
	\label{bdrate}
\end{figure}

In order to evaluate the proposed framework, we test HEVC sequences with a wide bitrate range controlled by Constant QP (CQP). 
For the conventional coding chain, the QP values are set between 28 and 48.
And in order to create Rate-Distortion (RD) curves for the MultiRC and MixedRC that are located in the same bitrate
range, QPs from 20 to 40 are used to compress the 2× down-sampled test sequences. 
For MultiRC, we use EDVR (2×)\cite{wang2019edvr} at the decoder side. 
For clarity, the RD curves of \textit{Traffic} and \textit{ParkScene} sequences are drawn exemplary in Fig.~\ref{bdrate} with PSNR as quality metric. 
And overall BD-rates are reported in Table~\ref{tab_BD}, with PSNR and Video Multimethod Assessment Fusion (VMAF) \cite{li2018vmaf} as quality metrics.

Experimental results show that both MixedRC and MultiRC chains surpass results of the conventional HEVC up to a certain bitrate around 7000 kbit/s, as well as MixedRC raise a curve at the upper left of the conventional and the MultiRC. 
Moreover, the proposed MixedRC shows superior results especially with VMAF as quality metric.

\begin{table}[]
	\centering
	\footnotesize
	\begin{tabular}{lllll}
	\hline
		\textbf{base \& anchor codec} & \textbf{HEVC sequences} & \textbf{PSNR} &  & \textbf{VMAF} \\
		\hline
		Conventional HEVC             & Class A                 & -14.43\%      &  & -25.20\%      \\
		& Class B                 & -16.35\%      &  & -26.77\%      \\
		& Class E                 & ~~-8.35\%       &  & -22.91\%      \\
		\hline
		MultiRC (with EDVR)           & Class A                 & ~~-7.08\%       &  & -18.20\%      \\
		& Class B                 & ~~-6.38\%       &  & -16.09\%      \\
		& Class E                 & ~~-3.91\%       &  & ~~-7.53\%      \\
	\hline
	\end{tabular}
\vspace{-6pt}
\caption{Coding performance comparison of MixedRC over HEVC anchor (HM) and MultiRC.}
\label{tab_BD}
\vspace{-8pt}
\end{table}

\begin{table*}[t]
	\vspace{-10pt}
	\centering
	\footnotesize
	\begin{tabular}{llllllllllllllll}
		\toprule
		\textbf{Codec Chain} & \textbf{Restoration Methods} &  & \multicolumn{5}{c}{\textbf{540p-to-1080p (2×)}}           &  &  &  & \multicolumn{5}{c}{\textbf{270p-to-1080p (4×)}} \\
		\hline
		& &  & \multicolumn{1}{c}{SSIM↑} &  & PSNR↑ &  & LPIPS↓ &  &  &  & SSIM↑   &    & PSNR↑   &    & LPIPS↓   \\
		\hline
		MultiRC& Bicubic        &  & 0.933                     &  & 30.11 &  & 0.195  &  &  &  & 0.912  &   & 27.83  &  & 0.312  \\
		MultiRC& RCAN\cite{zhang2018image}           &  & 0.939                     &  & 30.96 &  & 0.176  &  &  &  & 0.916  &   & 28.35  &  & 0.268  \\
		MultiRC& EDVR\cite{wang2019edvr}           &  & \underline{0.944}                     &  & 31.80 &  & 0.159  &  &  &  &  \underline{0.925}  &   &  \underline{29.51}  &  & 0.220  \\
		MixedRC& SRNTT\cite{zhang2019image}          &  & 0.940                     &  & 31.34 &  &  \underline{0.142}  &  &  &  & 0.921  &   & 29.29  &  &  \underline{0.181}  \\
		\hline
		MultiRC& MF-CNN\cite{yang2018multi} + Bicubic &  & 0.941                     &  & 31.65 &  & 0.163  &  &  &  & 0.916  &   & 28.21  &  & 0.283  \\
		MultiRC& MF-CNN\cite{yang2018multi} + EDVR\cite{wang2019edvr}    &  &  0.942                    &  &  \underline{31.97} &  & 0.156  &  &  &  & 0.922  &   & 29.36  &  & 0.242  \\
		\hline
		MixedRC& R3N(Ours)      &  & \textbf{0.951}                     &  & \textbf{32.49} &  & \textbf{0.138}  &  &  &  & \textbf{0.935}  &   & \textbf{30.58}  &  & \textbf{0.177} 
		\\
		\bottomrule
	\end{tabular}
	\caption{Quantitative results of coding chains with corresponding SR and CAR methods, for video restoration on our collected VT-100 (2× and 4×). Best and second best results are highlighted and underlined, and note that a lower LPIPS score indicates better image quality.}
	\vspace{-5pt}
	\label{metrics}
\end{table*}

\begin{figure*}
	\centering
	\subfigure[Ref (upper) \& LR]{
		\begin{minipage}[b]{0.15\linewidth}
			\includegraphics[width=1\linewidth]{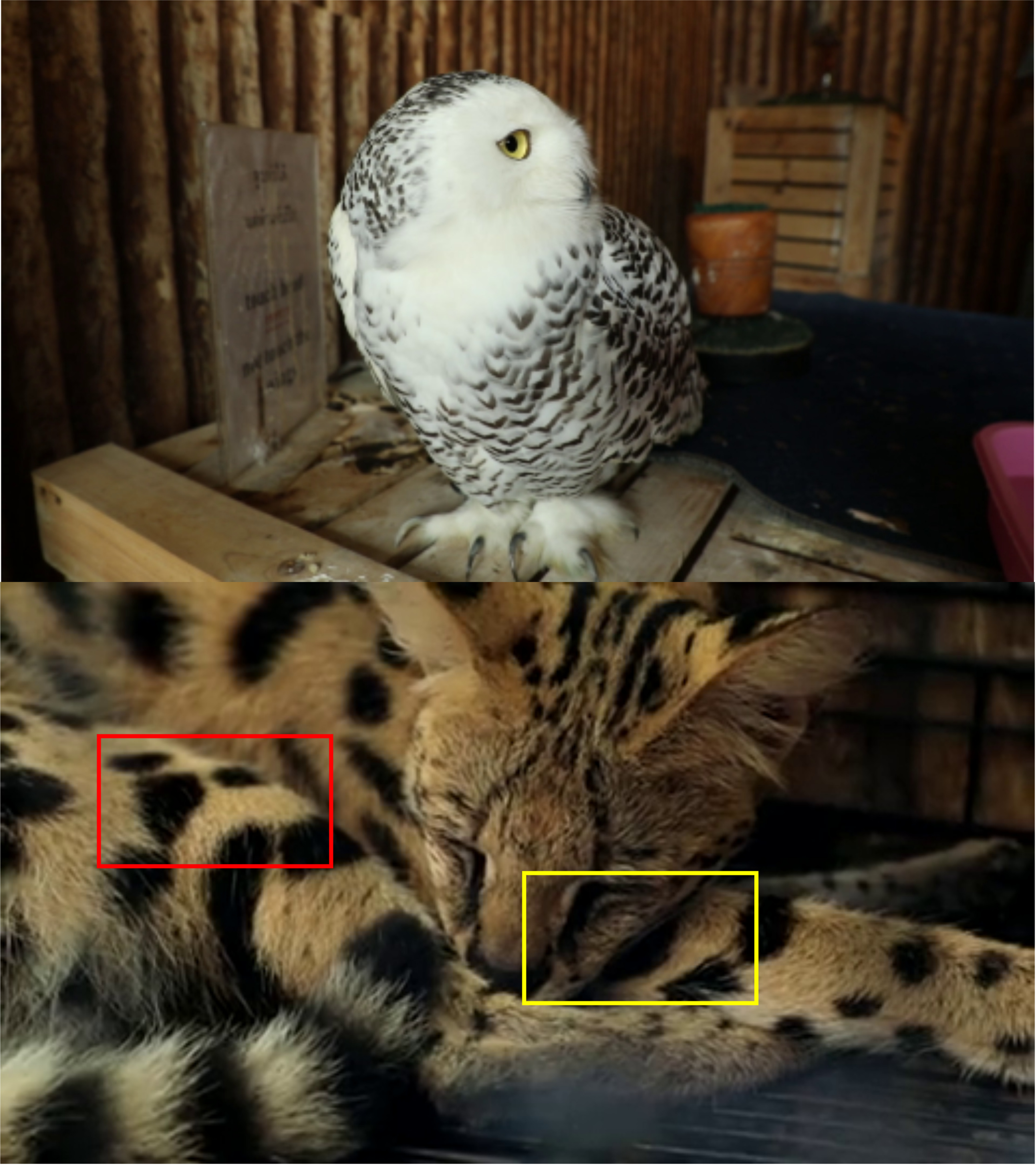}
	\end{minipage}}
	\subfigure[(a) HR]{
		\begin{minipage}[b]{0.15\linewidth}
			\includegraphics[width=1\linewidth]{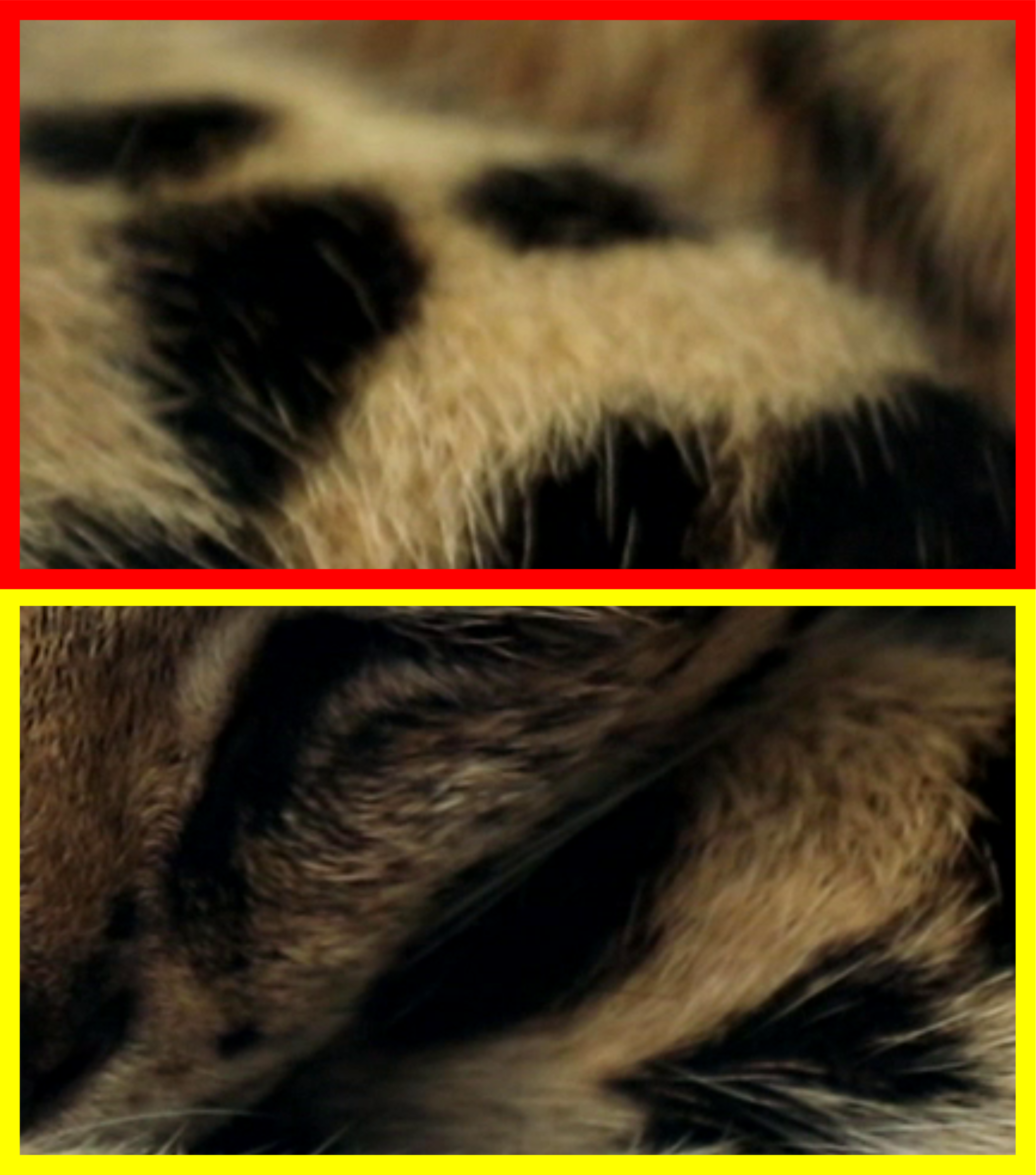}
	\end{minipage}}
	\subfigure[(b) Bicubic]{
		\begin{minipage}[b]{0.15\linewidth}
			\includegraphics[width=1\linewidth]{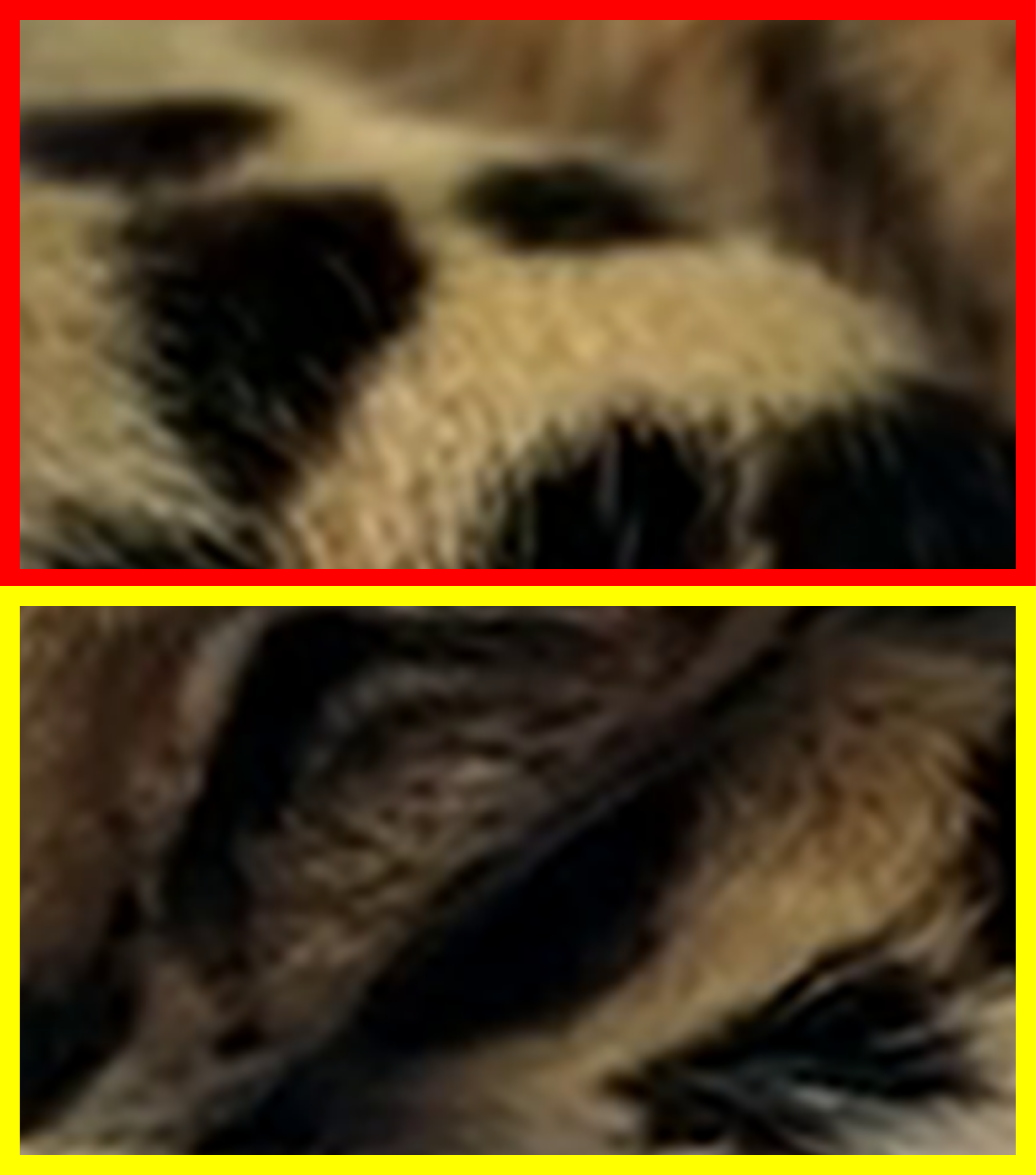}
	\end{minipage}}
	\subfigure[(c) EDVR]{
		\begin{minipage}[b]{0.15\linewidth}
			\includegraphics[width=1\linewidth]{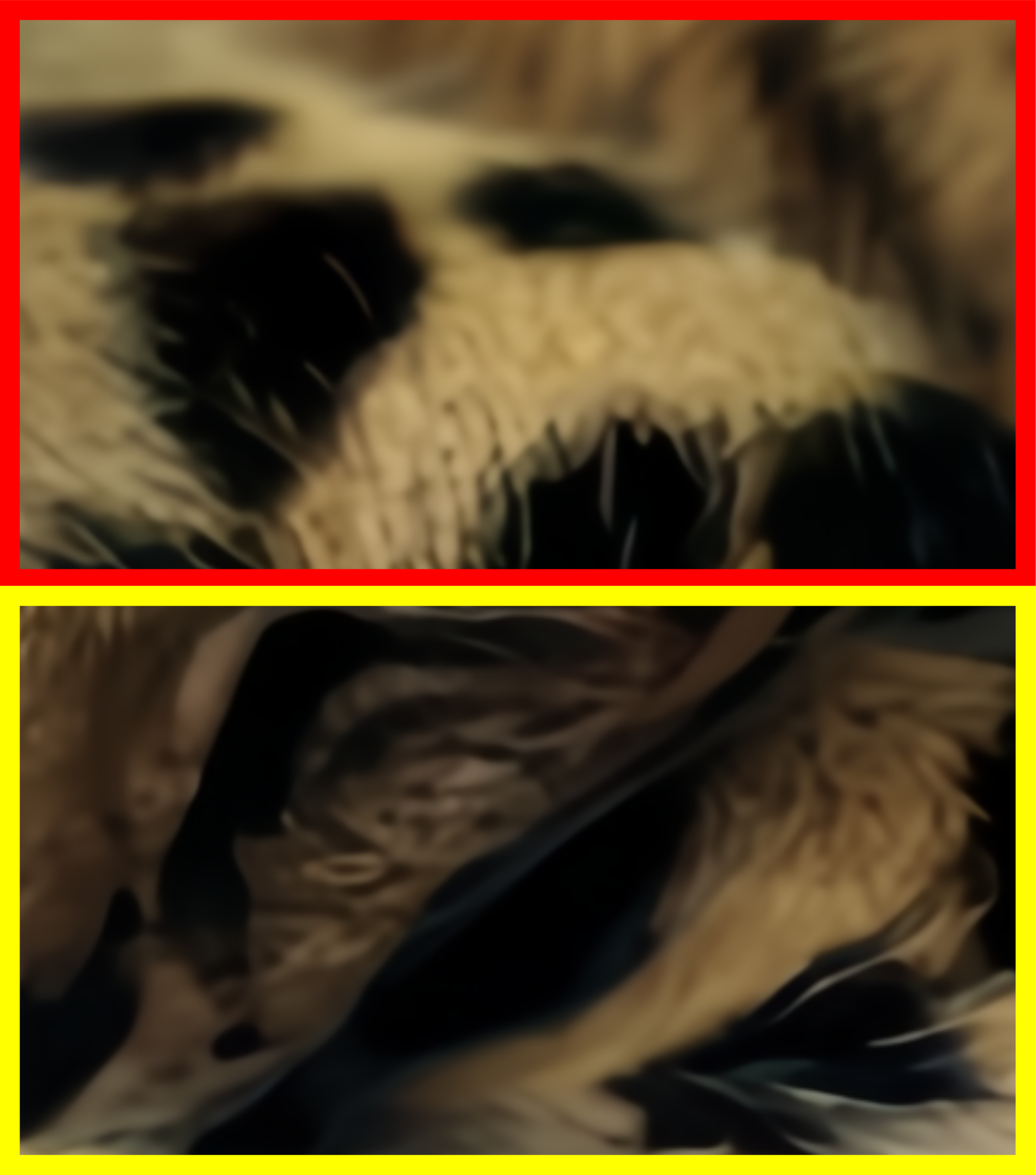}
	\end{minipage}}
	\subfigure[(d) SRNTT]{
		\begin{minipage}[b]{0.15\linewidth}
			\includegraphics[width=1\linewidth]{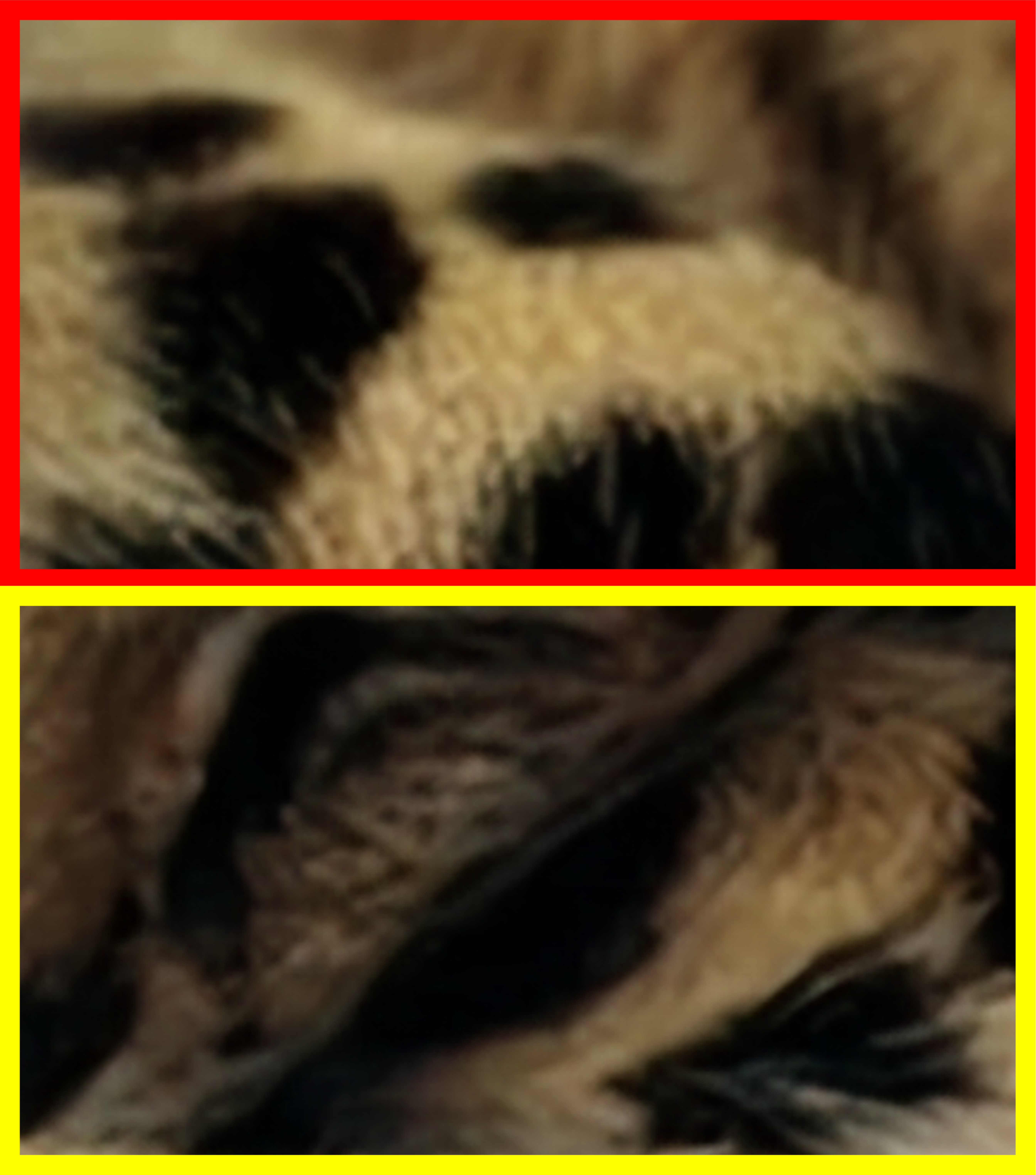}
	\end{minipage}}
	\subfigure[(e) R3N]{
		\begin{minipage}[b]{0.15\linewidth}
			\includegraphics[width=1\linewidth]{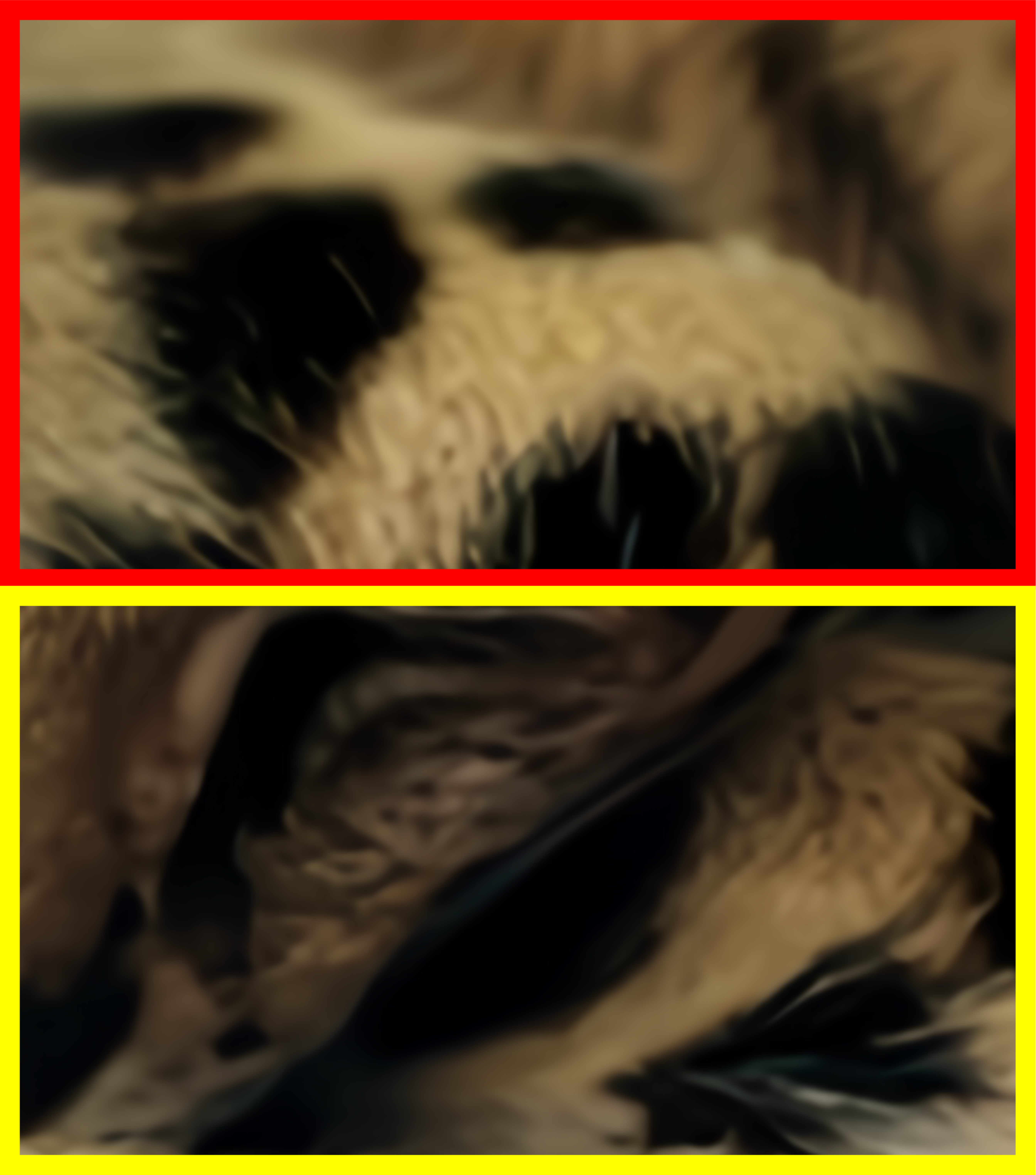}
	\end{minipage}}
	\subfigure[Ref (upper) \& LR]{
		\begin{minipage}[b]{0.15\linewidth}
			\includegraphics[width=1\linewidth]{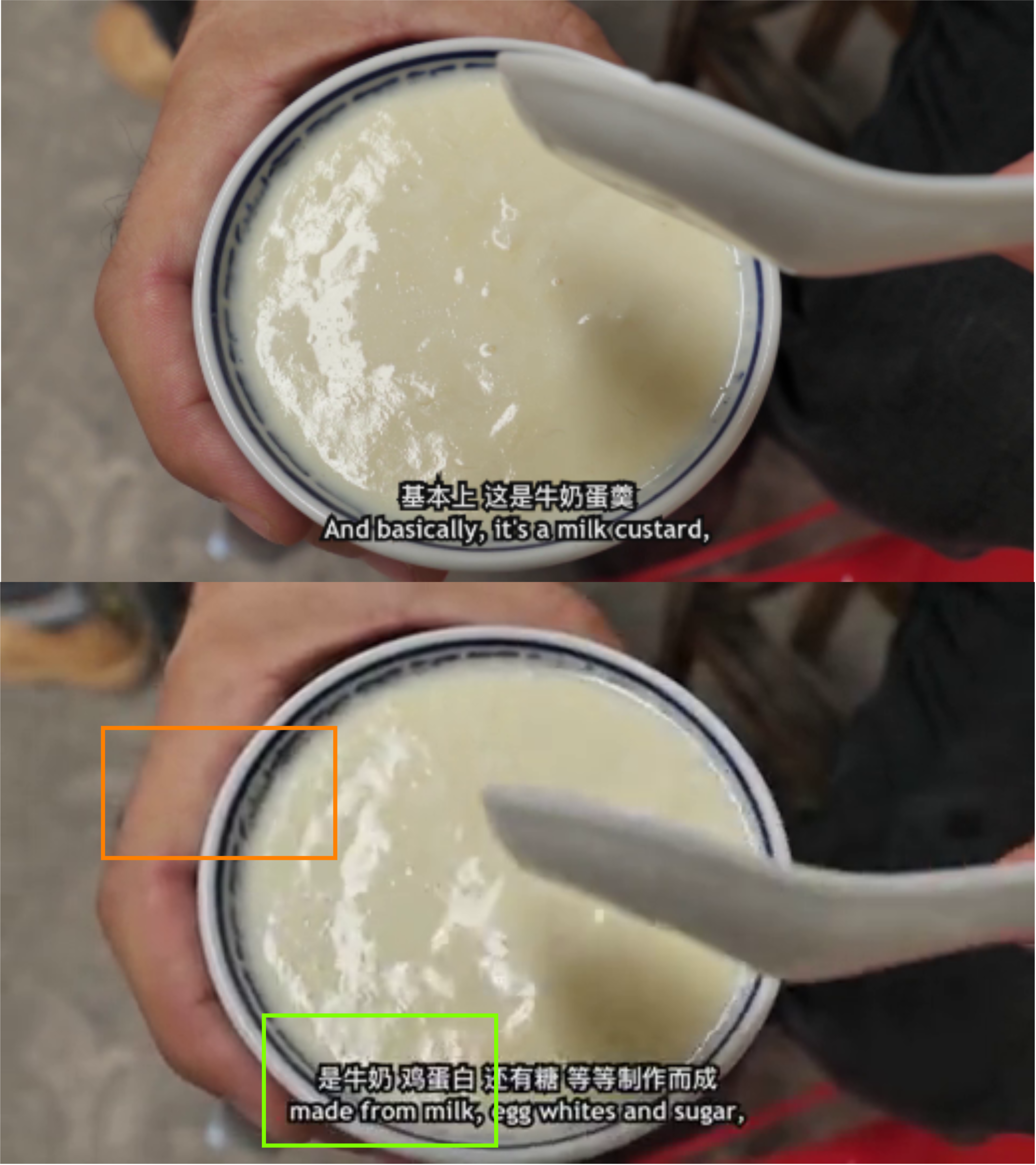}	
	\end{minipage}}
	\subfigure[(f) HR]{
		\begin{minipage}[b]{0.15\linewidth}
			\includegraphics[width=1\linewidth]{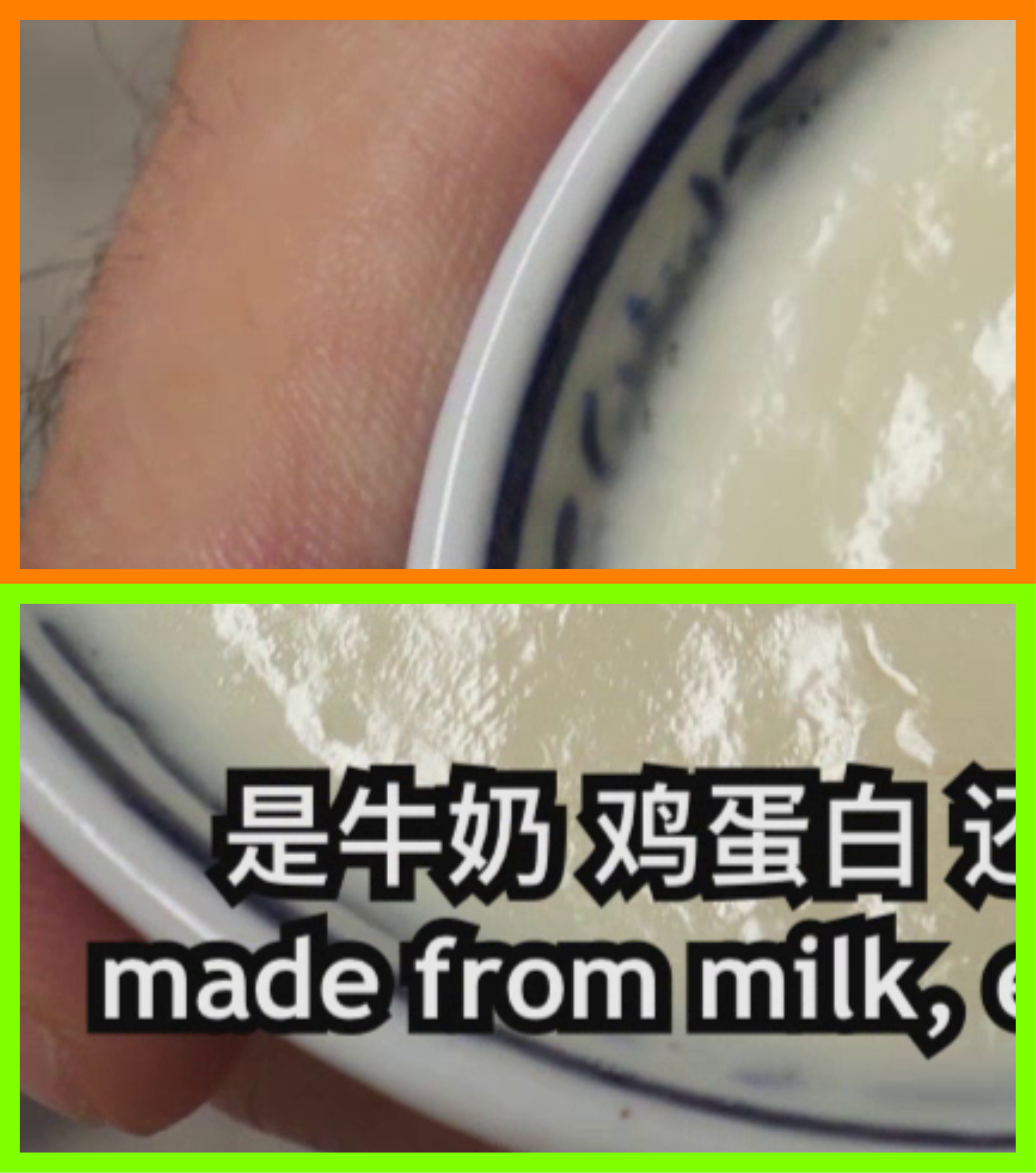}
	\end{minipage}}	
	\subfigure[(g) Bicubic]{
		\begin{minipage}[b]{0.15\linewidth}
			\includegraphics[width=1\linewidth]{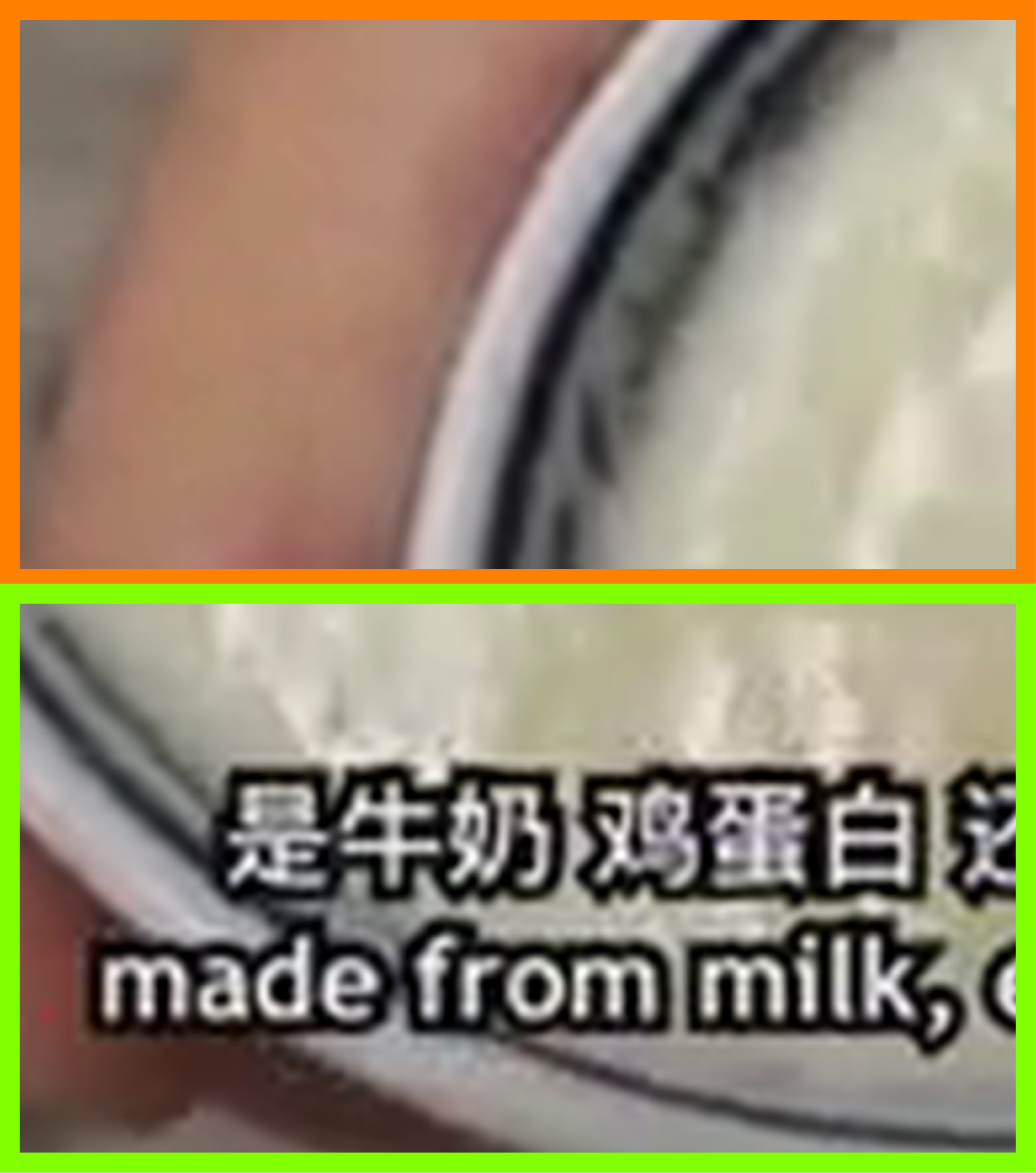}
	\end{minipage}}
	\subfigure[(h) EDVR]{
		\begin{minipage}[b]{0.15\linewidth}
			\includegraphics[width=1\linewidth]{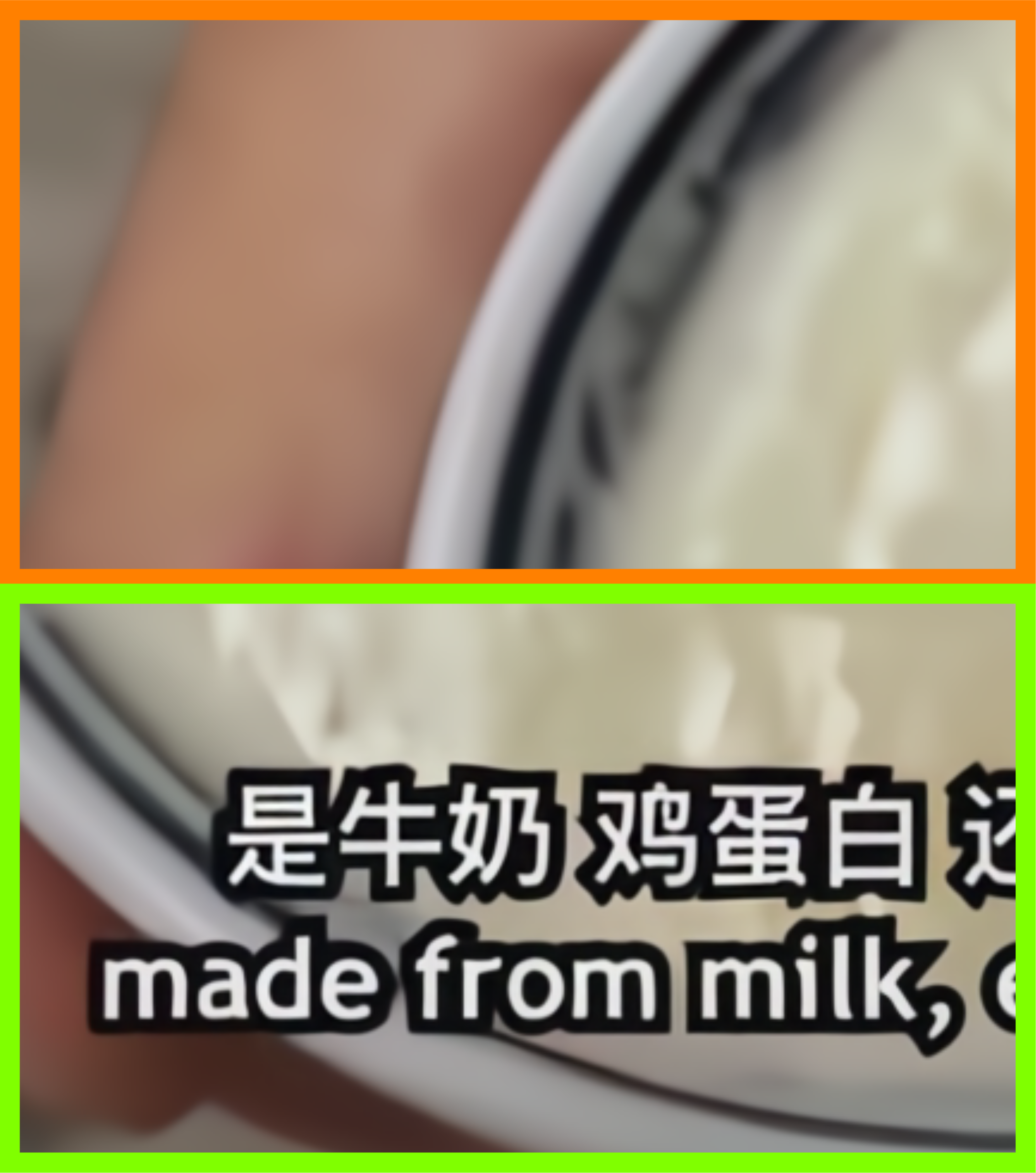}
	\end{minipage}}
	\subfigure[(i) SRNTT]{
		\begin{minipage}[b]{0.15\linewidth}
			\includegraphics[width=1\linewidth]{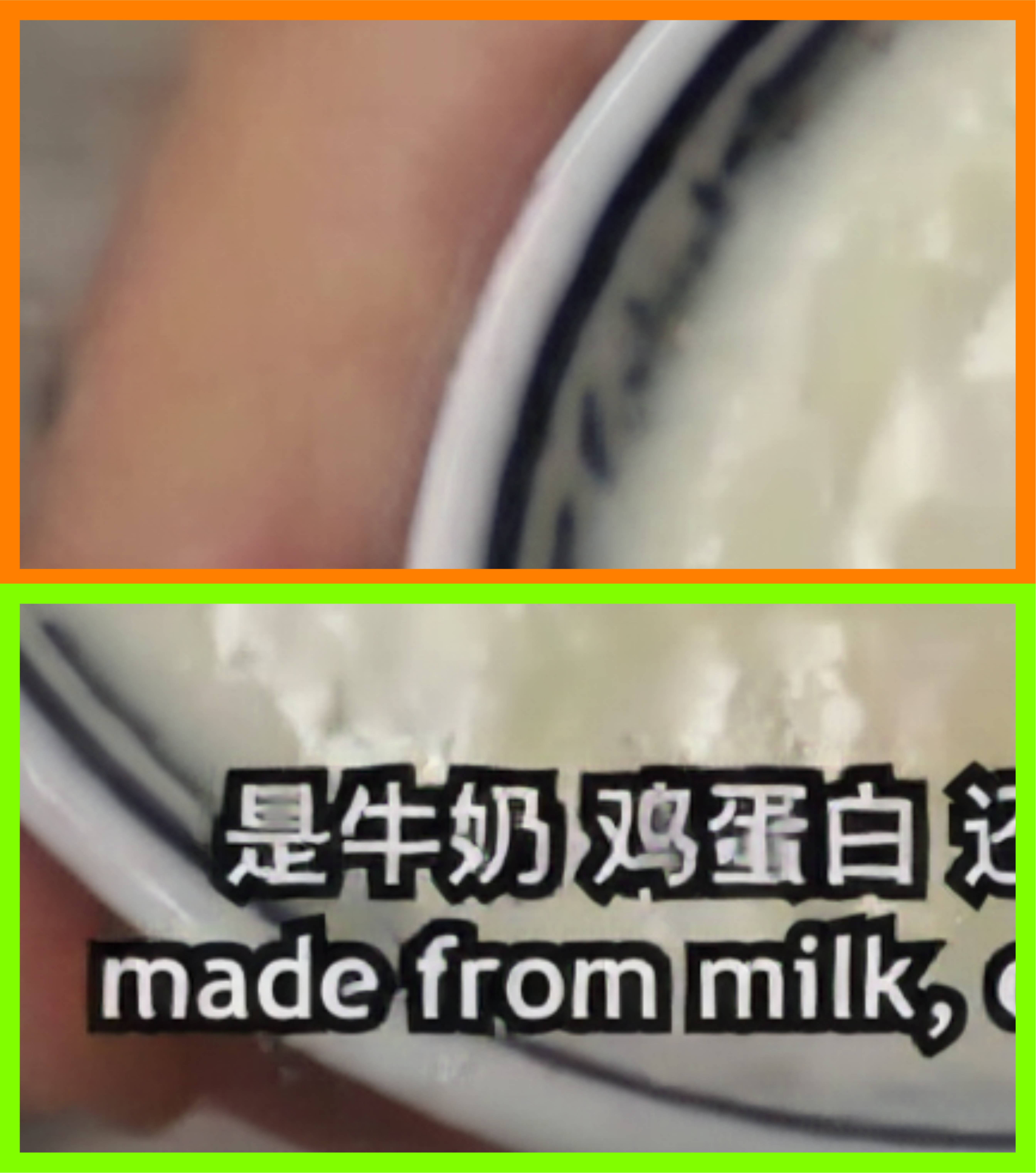}
	\end{minipage}}
	\subfigure[(j) R3N]{
		\begin{minipage}[b]{0.15\linewidth}
			\includegraphics[width=1\linewidth]{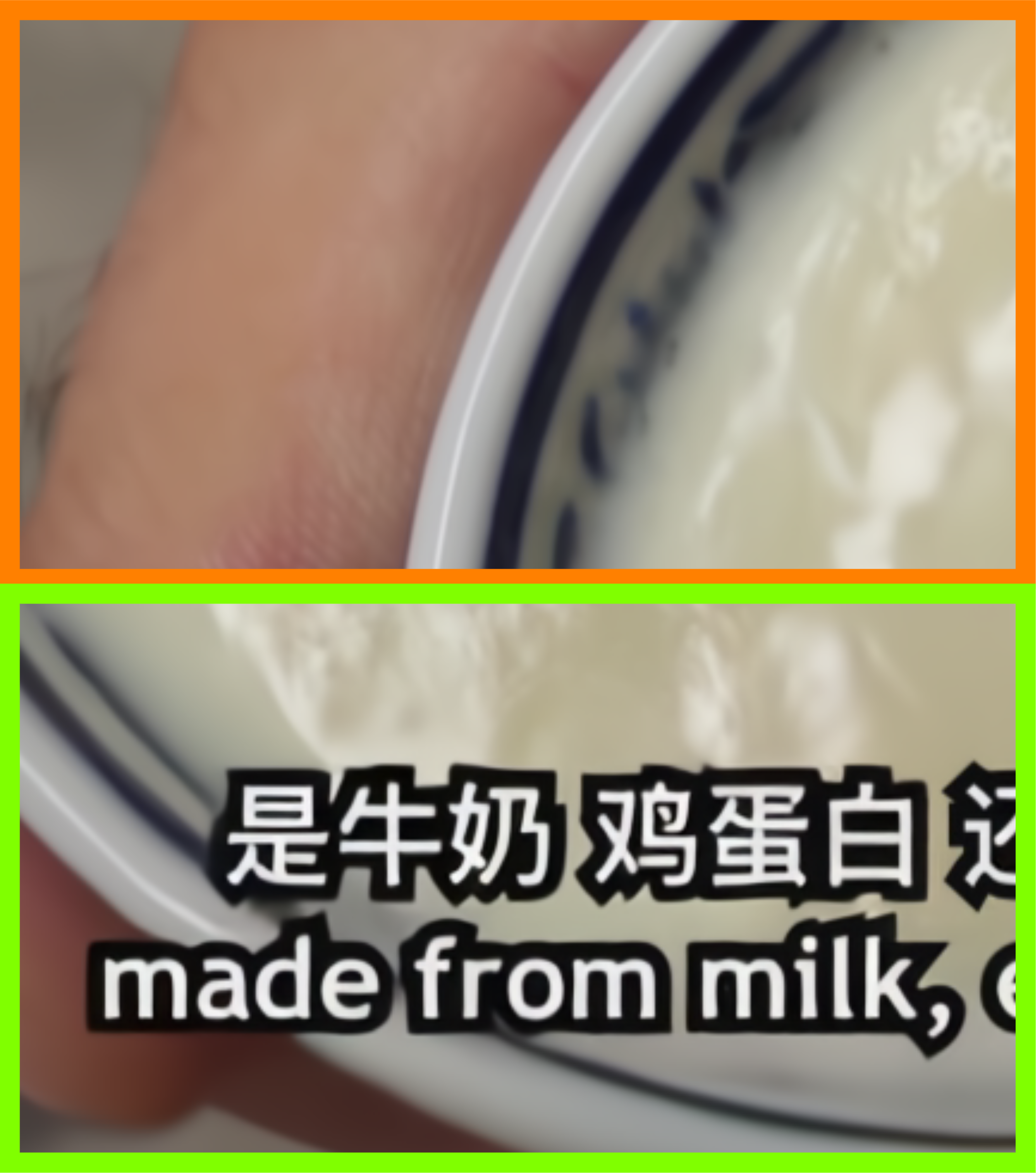}
	\end{minipage}}
	
	% 下 [edvr] PSNR:33.95, SSIM:0.9744, LPIPS:0.1357
	% 下 [r3n]  PSNR:32.72, SSIM:0.9760, LPIPS:0.1135
	
	% 上 [edvr] PSNR:29.79, SSIM:0.8996, LPIPS:0.2815
	% 上 [r3n] PSNR:29.67, SSIM:0.8975, LPIPS:0.2853
	\vspace{-10pt}
	\caption{R3N (ours) is compared to EDVR \cite{wang2019edvr} (a state-of-the-art video SR method) and SRNTT \cite{zhang2019image} (a state-of-the-art Ref-SR method). In the first row, the Ref image has irrelevant content to the LR input.
		In the second row, the Ref image and the LR input are in the same scene. }
	\label{visual}
\end{figure*}
\subsection{Network Settings}
The deformable alignment module adopts one offset generator and three offset refiner, and each offset refiner contains two Incep-HDC blocks. 
The network takes three consecutive LR frames as inputs in 2× model, and five consecutive frames in 4× model. As we observed through experimental results that the worse compression quality, the less complementary information in adjacent decoded frames.
% We use video clips from VT-1K to train a 2× model and a 4× model. 

We train our model with Adam optimizer by setting $\beta_{1} = 0.9$ and $\beta_{2} = 0.999$. The learning rate is initialized as $10^{-4}$ and then decreases to half whenever the validation loss stops decreasing for more than 5 epochs. 
Mini-batch size is set to 16. In each training mini-batch, patches randomly cropped with size 192 × 192 are used. By analyzing the dataset, we conclude that a large patch size is necessary for the Ref-SR task. 
And we train R3N following an intuitive idea of ``easy-hard transfer", which is discussed in Sec.~\ref{Ablation}.
We implement our models with the PyTorch framework and train them using 8 NVIDIA Tesla P40 GPUs.

\subsection{Network Performance}
We choose a few representative methods for comparisons: RCAN \cite{zhang2018image}, EDVR \cite{wang2019edvr}, SRNTT \cite{zhang2019image} which demonstrate different network architectures for single image SR, video SR and Ref-SR, and MF-CNN\cite{2019MFQE} which is designed to enhance the quality of compressed video. 
The structure of SR methods have also been proven to achieve good performance in other general low-level tasks. 
Therefore, we retrained these models to perform SR and CAR jointly.
It also because, to our knowledge, that this study is the first attempt to explore effectiveness of DCNN in joint SR-CAR problem in literature.
And for a fair comparison, we use the same data processing method and training data from VSRE-set. 
Since MF-CNN is not a SR model, we down-sampled ground-truth as the label for training it. And we enable memory-efficient forward so that SRNTT can be tested on HD video sequences.

To quantitatively evaluate the proposed method, we use the standard Structural Similarity (SSIM), Peak Signal To Noise Ratio (PSNR) and Learned Perceptual Image Patch Similarity (LPIPS) \cite{Zhang_2018_CVPR}. 
The results are reported in Table~\ref{metrics}. On VT-100, R3N and SRNTT achieve the best and second performance in terms of LPIPS. One can clearly see that R3N outperforms other methods by a large margin in terms of PSNR, SSIM and LPIPS.

Moreover, visual quality comparisons are presented in Fig.~\ref{visual}. Our proposed method recovers more high-fidelity visual information compared to others, especially in Fig.~\ref{visual} (j), where the skin texture is only restored by R3N. By contrast, SRNTT generates more twisty artifacts than ours. Our method could still achieve state-of-the-art SR performance when scene switching occurs, shown in Fig.~\ref{visual} (e), demonstrating the robustness of R3N. Furthermore, R3N is superior to EDVR and SRNTT on the implementation speed, shown in Table~\ref{speed}.

\begin{table}[]
	\centering
	\footnotesize
	\begin{tabular}{llccc}
		\toprule
		Restoration Methods & RCAN                      & EDVR  & SRNTT   & R3N (Ours) \\
		\hline
		Times (second)      & \multicolumn{1}{c}{0.321} & 0.694 & 30.163 & 0.473    \\
		\bottomrule
	\end{tabular}
	\vspace{-5pt}
	\caption{Test the speeds of restoration methods (reconstructing a frame from 270p to 1080p using a GPU of NVIDIA Tesla V100).}
	\label{speed}
\end{table}

\section{Ablation Study}
\label{Ablation}
We study the effects of each component in the proposed method, as reported in Table~\ref{tab_ablation}.

\begin{figure}[]
	\centering
	\subfigure[(a) Ref \& LR]{
		\begin{minipage}[b]{0.45\linewidth}
			\includegraphics[width=1\linewidth]{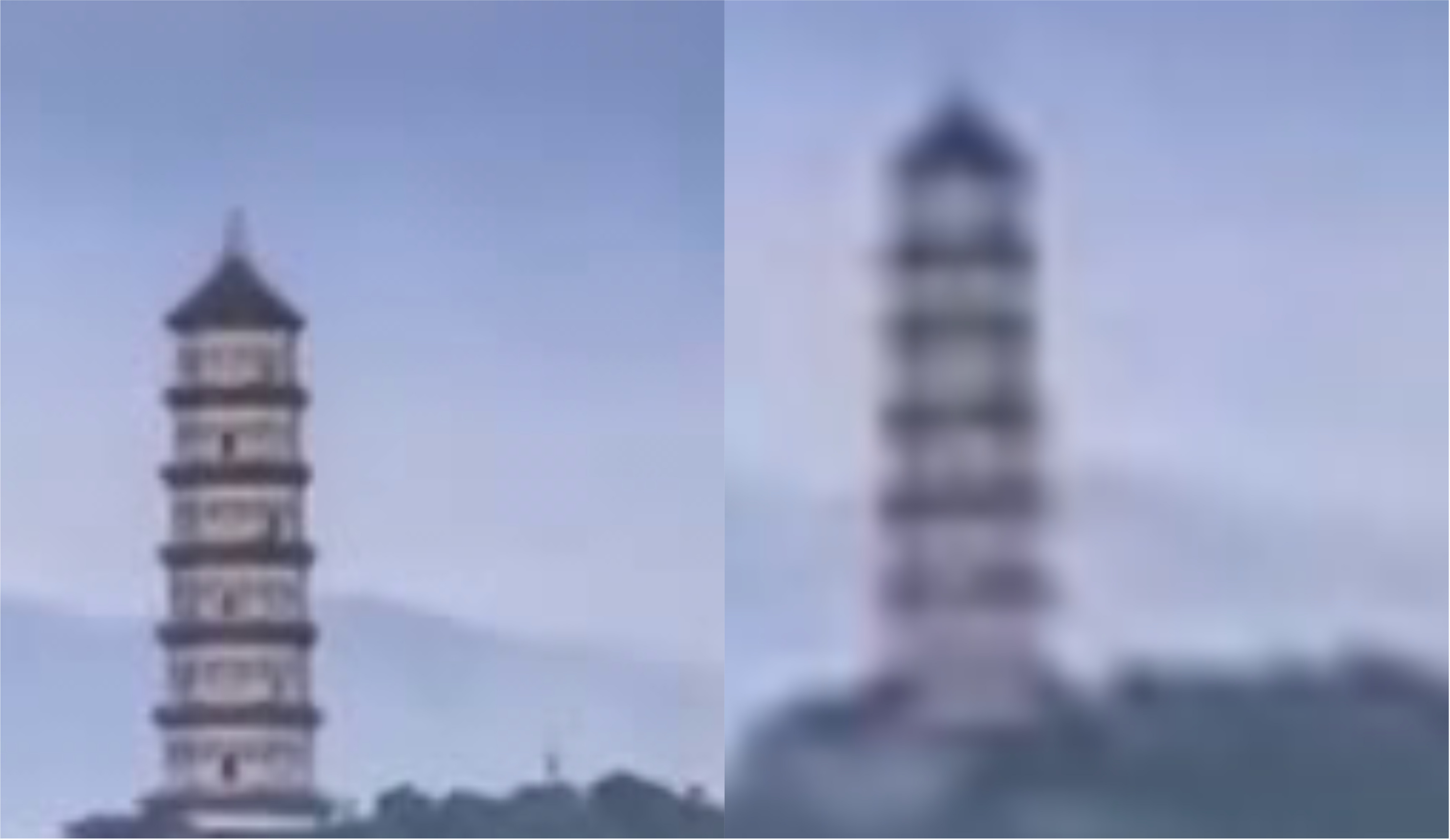}
	\end{minipage}}
	\subfigure[(b) w/o refined-offset alignment]{
		\begin{minipage}[b]{0.45\linewidth}
			\includegraphics[width=1\linewidth]{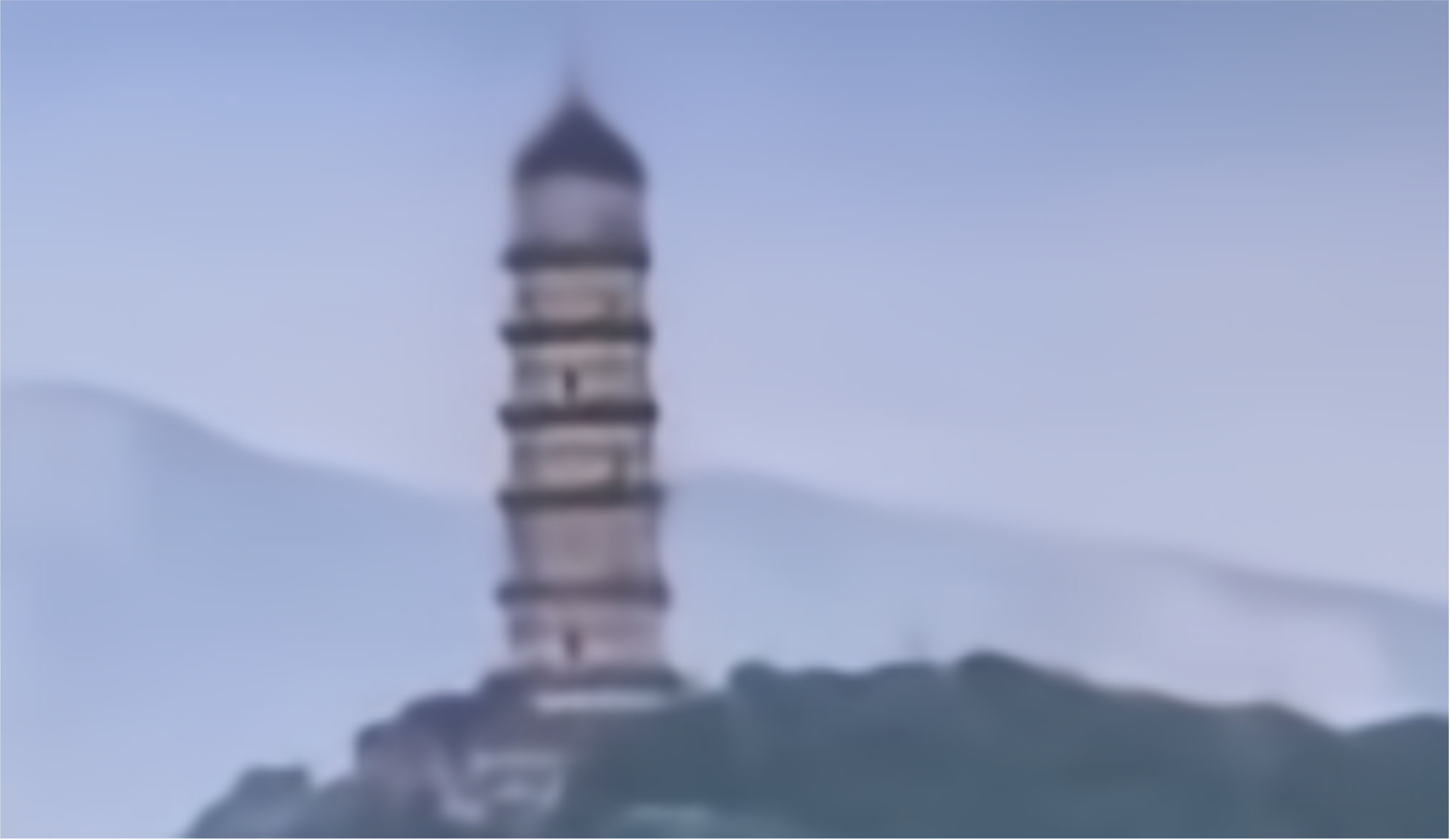}
	\end{minipage}}
	\subfigure[(c) w/o $\mathcal{L}_{distan}$]{
		\begin{minipage}[b]{0.45\linewidth}
			\includegraphics[width=1\linewidth]{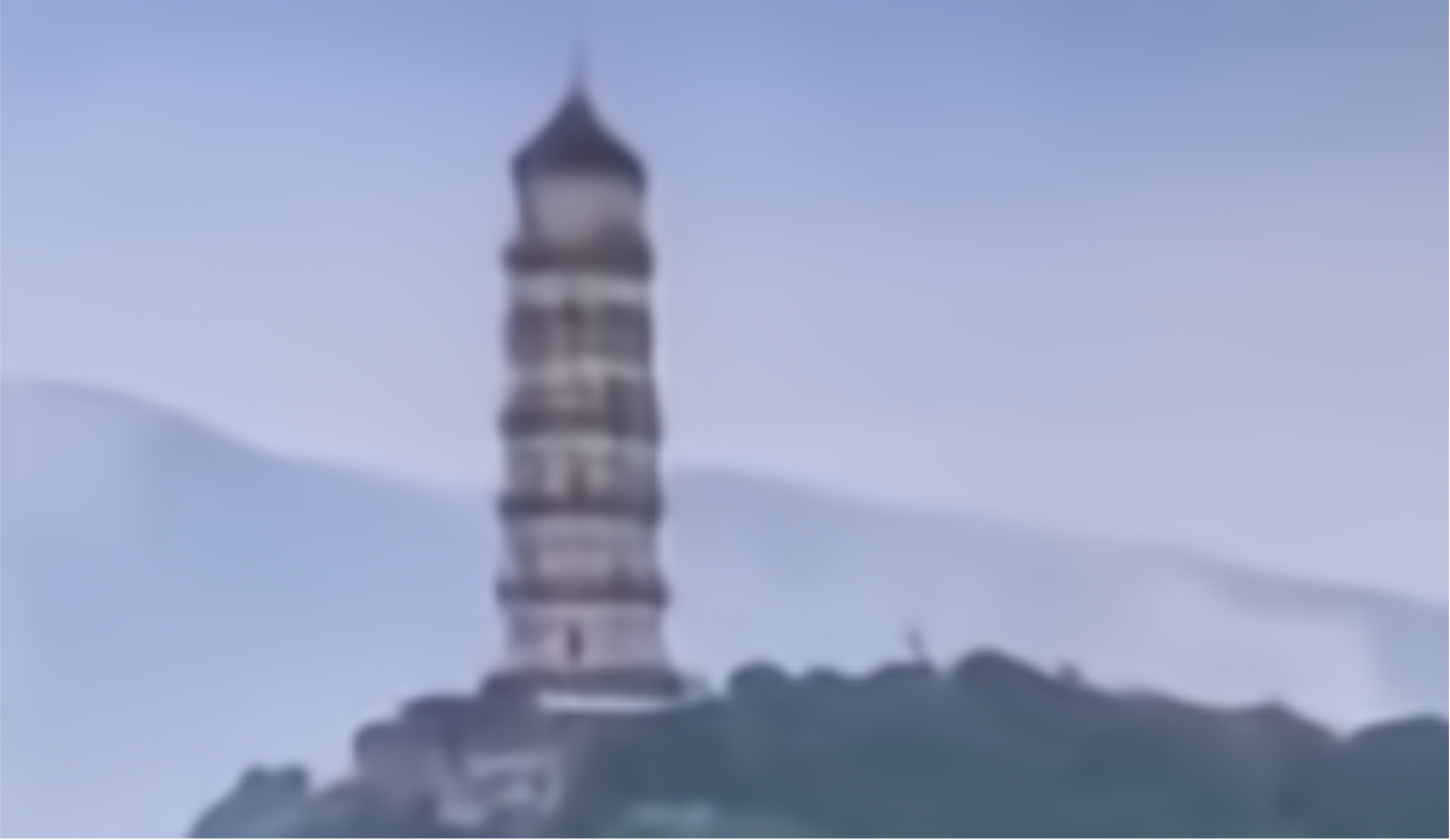}
	\end{minipage}}
	\subfigure[(d) the proposed ]{
		\begin{minipage}[b]{0.45\linewidth}
			\includegraphics[width=1\linewidth]{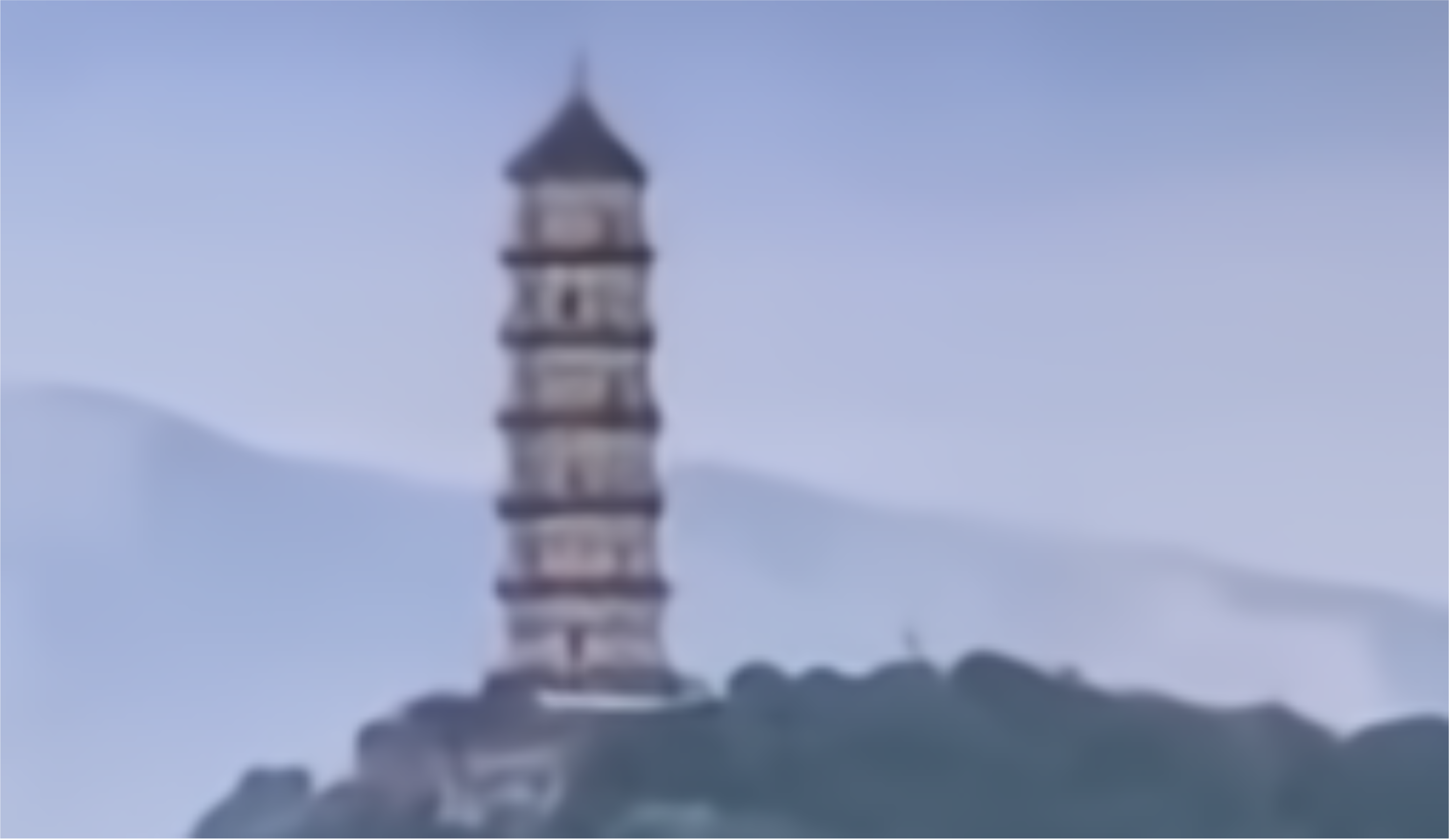}
	\end{minipage}}
	\vspace{-10pt}
	\caption{Visual comparison for ablation testing. Note that there are significant motion between Ref and LR frames, which could be seen clearly by embedding their patches next to each other in (a). }
\end{figure}

\begin{table}[]
	\centering
	\footnotesize
	\begin{tabular}{l|l|l|l|l|l}
		\hline
		Method                                 & (a) & (b) & (c) & (d) & (e) \\
		\hline
		\hline
		Stacking DeformConv & \checkmark  &     &     &     &     \\
		\hline
		Refined-offset    &     & \checkmark  & \checkmark  & \checkmark  & \checkmark  \\
		\hline
		Channel attention module               &   & \checkmark  &     &     &     \\
		\hline
		Spatial attention module             &  \checkmark   &     & \checkmark  & \checkmark  & \checkmark  \\
		\hline
		Disentangled loss                      &     &     &     & \checkmark  & \checkmark  \\
		\hline
		Easy-hard transfer                     &     &     &     &     & \checkmark  \\
		\hline
		\hline
		VT-100 (4×)                              & 29.46 & 29.87 & 30.12 & 30.31 & 30.58 \\
		\hline  
	\end{tabular}
	\vspace{-5pt}
	\caption{Ablation study on the proposed methods. Our results are validated by the PSNR score.}
	\label{tab_ablation}
\end{table}

\noindent{\bfseries Effectiveness of Refined-offset Deformable Alignment.}
As illustrated in Table~\ref{tab_ablation}, we introduce two baselines: (a) and (c), where the model (a) stacks three layers of deformable convolution.  For fairness, we set the same total number of Incep-HDC blocks as 16 in (a) and (c). 
As expected, R3N has at least an average PSNR gain of 0.6dB over stacking deformable convolution layers due to the residual learning manner. Visual comparisons are provided in Fig.~\ref{Ablation} (b) and (d). 

\noindent{\bfseries Effectiveness of Disentangled Loss.}
To demonstrate the effectiveness of disentangled loss $\mathcal{L}_{distan}$, we train one of our models with  $ \ell_1$ only and another same model with $\mathcal{L}_{distan}$ and $ \ell_1$, and here p = 1 in Eq.~(\ref{eq:loss}). 
In Fig.~\ref{Ablation} (c), we can see result without $\mathcal{L}_{distan}$ would suffer from some erroneous texture artifacts. 
Such observations futher prove the effectiveness of our proposed disentangled loss.

\noindent{\bfseries Effect of Attention Module.}
When we compare the results of (b) and (c) in Table~\ref{tab_ablation}, we find that Incep-HDC with spatial attention would perform better than those with channel attention. 
The implementation of spatial and channel attention is in line with RFANet \cite{liu2020residual} and RCAN \cite{zhang2018image}.

\noindent{\bfseries Effect of Transfer Learning.}
Transfer learning in deep models provides a good starting point that could help train a better network with higher convergence speed. 
In frame alignment of Ref-SR and video SR, it is observed that using training data with large motion would encounter the problem of convergence. We are also met with this difficulty during the training process of R3N. 
To this end, we conduct the transfer learning settings as follows:
we trained a model whose Ref is selected randomly between the $I^{Ref}_{t-d}$ and $I^{Ref}_{t+d}$, where \textit{d}=8 at first.
Then we reuse the features learned in the former trained model to initialize a harder network, where \textit{d} is increased to 16, 24 and 32 gradually.
Results in Table~\ref{tab_ablation} show that easy-hard transfer can bring a gain of 0.27dB, compared with the model trained directly with \textit{d}=32.

\section{Conclusion}
\label{discussion}
This paper explains that advancement in deep convolutional network opens up expanded design space and has far-reaching implications in the video transmission ecosystem. 
First, we introduced the mixed-resolution video coding which not only improves the coding efficiency, but also leaves more improvement room for restoration algorithms.
However, the picture quality loss in video transmission mainly comes from resolution loss and compression artifacts.
Existing SR algorithms suffer a huge peformance degradation when encountering this intricate task.
So it is of particular interest to bring a joint SR-CAR solution.

Aiming at this problem, we embrace and carry on improvement on Ref-SR methods. 
We propose a refined-offset deformable alignment module, at the mean time, we propose a disentangled loss to distinguish different components in high-frequency regions.
It is worth mentioning that the proposed loss also has potential for the denoising task. We will leave them to further work.

%\noindent{\bfseries{Acknowledgment.}}
%Thanks to National Natural Science Foundation of China 61672063, Shenzhen Research Projects of JCYJ20180503182128089 and 201806080921419290.

% \bibliographystyle{ieee_fullname}
{\small
	\bibliographystyle{ieee_fullname}
	\bibliography{egbib}
}

\end{document}